\documentclass{article}

\usepackage{microtype}
\usepackage{graphicx}
\usepackage{booktabs} 

\usepackage{enumitem}
\usepackage{amsmath}
\usepackage{amsfonts}
\usepackage{dsfont}
\usepackage{standard_definitions}
\usepackage{amssymb}
\usepackage{pifont}
\usepackage{array}
\usepackage{xspace}
\usepackage{nicefrac}
\usepackage{xcolor}
\usepackage[font=small,skip=0pt]{caption}
\usepackage{subcaption}
\usepackage{afterpage}
\usepackage{todonotes}

\usetikzlibrary{shapes,snakes}
\usetikzlibrary{shapes.geometric}

\usepackage{algorithmicx,algorithm}
\usepackage[noend]{algpseudocode}

\usepackage{hyperref}

\setlist[itemize]{noitemsep, topsep=0pt}

\newcommand{\pitheta}{\pi_{\theta}}
\newcommand{\pibeta}{\pi_{\beta}}
\newcommand{\piomega}{\pi_{\omega}}

\newcommand{\IL}{\textsc{IL}\xspace}
\newcommand{\RL}{\textsc{RL}\xspace}
\newcommand{\reinforce}{\textsc{Reinforce}\xspace}

\newcommand{\supp}{{\tt supp}}
\newcommand{\unf}{{\tt unf}}

\newcommand{\nav}{\textsc{Nav}\xspace}
\newcommand{\regex}{\textsc{Regex}\xspace}

\newtheorem{assumption}{Assumption}

\newcommand{\smin}{\sigma_{min}}

\definecolor{redbrown}{rgb}{0.522, 0.1, 0.18} 
\definecolor{darkgreen}{RGB}{0,118,0}
\definecolor{darkred}{RGB}{204,0,0}
\definecolor{mediumblue}{RGB}{0,102,204}
\definecolor{khanhcomment}{rgb}{0.76, 0.13, 0.28}

\newcommand{\mainsetup}{\textsc{Iliad}\xspace}
\newcommand{\mainalg}{\textsc{Adel}\xspace}
\newcommand{\epochmainalg}{\textsc{EpochAdel}\xspace}

\def\equationautorefname~#1\null{Eq~#1\null}
\renewcommand{\sectionautorefname}{\S\kern-0.2em}
\renewcommand{\subsectionautorefname}{\S\kern-0.2em}
\renewcommand{\subsubsectionautorefname}{\S\kern-0.2em}

\newcommand{\yellowRoute}{\tikz\draw[very thick,color={rgb,255:red,255;green,220;blue,0},scale=0.5](0,0) .. controls (0,0.5) and (0.5,0) .. (0.5,0.5);\xspace}
\newcommand{\redRoute}{\tikz\draw[very thick,color={rgb,255:red,255;green,0;blue,0},scale=0.5](0,0) .. controls (0,0.5) and (0.5,0) .. (0.5,0.5);\xspace}
\newcommand{\cyanRoute}{\tikz\draw[very thick,color={rgb,255:red,0;green,220;blue,220},scale=0.5](0,0) .. controls (0,0.5) and (0.5,0) .. (0.5,0.5);\xspace}

\usepackage[accepted]{icml2021}

\icmltitlerunning{Interactive Learning from Activity Description}

\begin{document}

\twocolumn[
\icmltitle{Interactive Learning from Activity Description}



\icmlsetsymbol{equal}{*}

\begin{icmlauthorlist}
\icmlauthor{Khanh Nguyen}{umd}
\icmlauthor{Dipendra Misra}{msr}
\icmlauthor{Robert Schapire}{msr}
\icmlauthor{Miro Dud\'ik}{msr}
\icmlauthor{Patrick Shafto}{rut}
\end{icmlauthorlist}

\icmlaffiliation{umd}{Department of Computer Science, University of Maryland, Maryland, USA}
\icmlaffiliation{msr}{Microsoft Research, New York, USA}
\icmlaffiliation{rut}{Rutgers University, New Jersey, USA}

\icmlcorrespondingauthor{Khanh Nguyen}{kxnguyen@umd.edu}

\icmlkeywords{Machine Learning Protocol, Interactive Learning, Reinforcement Learning, Imitation Learning, Learning from Language, Language Feedback, Interactive Learning from Activity Description, ICML}

\vskip 0.3in
]



\printAffiliationsAndNotice{}  

\begin{abstract}
We present a novel interactive learning protocol that enables training request-fulfilling agents by verbally describing their activities.
Unlike imitation learning (IL), our protocol allows the teaching agent to provide feedback in a language that is most appropriate for them.
Compared with reward in reinforcement learning (RL), the description feedback is richer and allows for improved sample complexity.
We develop a probabilistic framework and an algorithm that practically implements our protocol.
Empirical results in two challenging request-fulfilling problems demonstrate the strengths of our approach: compared with RL baselines, it is more sample-efficient; compared with IL baselines, it achieves competitive success rates 
without requiring the teaching agent to be able to demonstrate the desired behavior using the learning agent's actions.
Apart from empirical evaluation, we also provide theoretical guarantees for our algorithm under certain assumptions about the teacher and the environment.\looseness=-1

\end{abstract}

\section{Introduction}

The goal of a \emph{request-fulfilling} agent is to map a given
request in a situated environment to an execution that accomplishes the intent of the request~\cite{Winograd:72,chen2011learning,tellex-etal-2012-toward,artzi-etal-2013-semantic,Misra:17instructions,anderson2018vision, Chen19:touchdown,nguyen2019vnla,nguyen2019hanna,gaddy-klein-2019-pre}.
Request-fulfilling agents have been typically trained using \textit{non-verbal} interactive learning protocols such as imitation learning
($\IL$) which assumes labeled executions as feedback~\cite{mei2016listen,anderson2018vision,yao-etal-2020-imitation}, or reinforcement learning ($\RL$) which uses scalar rewards as feedback~\cite{chaplot2017gated,Hermann2017}. 
These protocols are suitable for training agents with pre-collected datasets or in simulators, but they do not lend themselves easily to training by human teachers that only possess domain knowledge, but might not be able to precisely define the reward function, or provide direct demonstrations.
To enable training by such teachers,
we introduce a verbal interactive learning protocol called $\mainsetup$: \textbf{I}nteractive \textbf{L}earning from \textbf{A}ctivity \textbf{D}escription, where feedback is limited to \textit{descriptions of activities}, in a language that is appropriate for a given teacher (e.g., a natural language for humans).

\begin{figure}[t!]
    \centering
     \includegraphics[width=\linewidth]{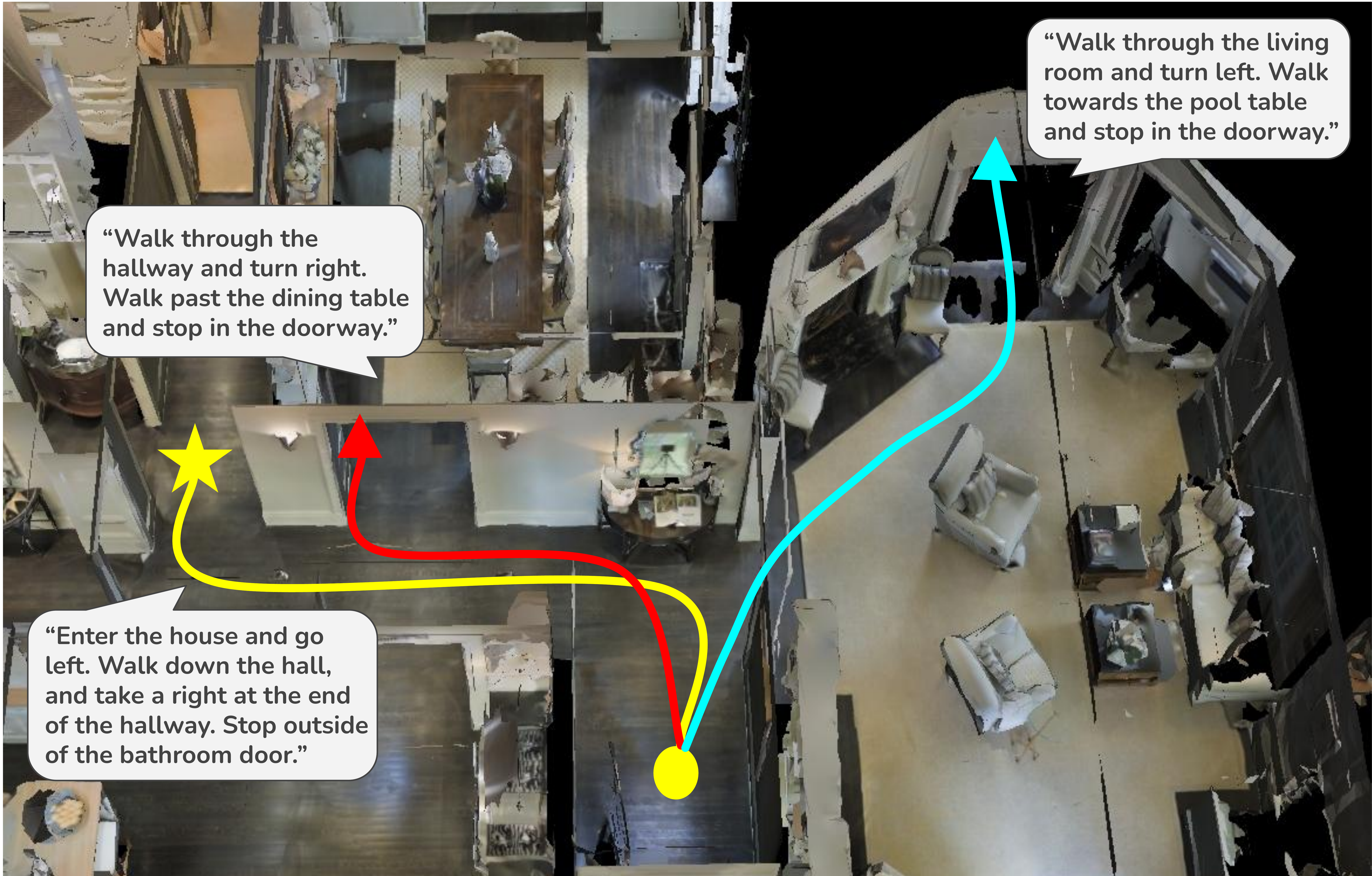}
    \caption[hello]{A real example of training an agent to fulfill a navigation request in a 3D environment \citep{anderson2018vision} using \mainalg, our implementation of the \mainsetup protocol. The agent receives a request \textit{``Enter the house..."} which implies the \yellowRoute path. Initially, it wanders far from the goal because it does not understand language. Its execution (\cyanRoute) is described as \textit{``Walk through the living room..."}. To ground language, the agent learns to generate the \cyanRoute path conditioned on the description. After a number of interactions, its execution (\redRoute) is closer to the optimal path. As this process iterates, the agent learns to ground diverse descriptions to executions and can execute requests more precisely.}
    \label{fig:iliad_example}
\end{figure}

\begin{table}[t!]
\centering
\scriptsize
\setlength{\tabcolsep}{3pt}
\caption{Trade-offs between the \textit{learning} effort of the agent and the teacher in three learning protocols. 
Each protocol employs a different medium for the teacher to convey feedback. 
If a medium is not natural to the teacher (e.g., \IL-style demonstration), it must learn to express feedback using that medium (\textit{teacher communication-learning effort}).
For example, in \IL, to provide demonstrations, the teacher must learn to control the agent to accomplish tasks. 
Similarly, if a medium is not natural to the agent (e.g., human language), it needs to learn to interpret feedback (\textit{agent communication-learning effort}). 
The agent also learns tasks from information decoded from feedback (\textit{agent task-learning effort}).
The qualitative claims about the ``agent learning effort'' column summarize our empirical findings about the learning efficiency of algorithms that implement these protocols (\autoref{tab:main}).}
\vspace{0.1cm}
\begin{tabular}{lccc}
    \toprule
     & & \multicolumn{2}{c}{Learning effort} \\ \cmidrule(lr){3-4} 
      & Feedback & Teacher & Agent  \\ 
     \multicolumn{1}{c}{Protocol}& medium & {\tiny(communication learning)} & {\tiny(comm. \& task learning)}   \\ 
     \midrule 
     \IL & Demonstration & \textcolor{darkred}{Highest} & \textcolor{darkgreen}{Lowest}   \\
     \RL & Scalar reward & \textcolor{darkgreen}{None} & \textcolor{darkred}{Highest} \\
     $\mainsetup$ & Description & \textcolor{darkgreen}{None} & \textcolor{mediumblue}{Medium}  \\
     \bottomrule 
\end{tabular}
\label{tab:tradeoff}
\end{table}

\autoref{fig:iliad_example} illustrates an example of training an agent using the \mainsetup protocol.
Learning proceeds in episodes of interaction between a learning agent and a teacher. In each episode, the agent is presented with a request, provided in the teacher's description language, and takes a sequence of actions in the environment to execute it. 
After an execution is completed, the teacher provides the agent with a description of the execution, in the same description language.
The agent then uses this feedback to update its policy.\looseness=-1

The agent receives \emph{no other} feedback such as ground-truth demonstration~\cite{mei2016listen}, scalar reward~\cite{Hermann2017}, or constraint~\cite{Miryoosefi2019constraints}. 
Essentially, \mainsetup presents a setting where task learning is enabled by \textit{grounded} language learning: the agent improves its request-fulfilling capability by exploring the description language and learning to ground the language to executions. 
This aspect distinguish \mainsetup from \IL or \RL, where task learning is made possible by imitating actions or maximizing rewards.

The \mainsetup protocol leaves two open problems: (a) \textit{the exploration problem}: how to generate executions that elicit useful descriptions from the teacher and (b) \textit{the grounding problem}: how to effectively ground descriptions to executions. 
We develop an algorithm named $\mainalg$: \textbf{A}ctivity-\textbf{D}escription \textbf{E}xplorative \textbf{L}earner that offers practical solutions to these problems.
For (a), we devise a semi-supervised execution sampling scheme that efficiently explores the description language space.
For (b), we employ maximum likelihood to learn a mapping from descriptions to executions.
We show that our algorithm can be viewed as density estimation, and prove its convergence in the contextual bandit setting~\citep{langford2008cb}, i.e., when the task horizon is 1.\looseness=-1

Our paper does \textit{not} argue for the primacy of one learning protocol over the others.
In fact, an important point we raise is that there are multiple, possibly competing metrics for comparing learning protocols. 
We focus on highlighting the \textit{complementary} advantages of \mainsetup against \IL and \RL (\autoref{tab:tradeoff}).
In all of these protocols, the agent and the teacher establish a communication channel that allows the teacher to encode feedback and send it to the agent.
At one extreme, \IL uses demonstration, an \textit{agent-specific} medium, to encode feedback, thus placing the burden of establishing the communication channel entirely on the teacher. 
Concretely, in standard interactive \IL (e.g., \citealp{ross2011reduction}), a demonstration can contain only actions in the agent's action space.
Therefore, this protocol implicitly assumes that the teacher must be familiar with the agent's control interface. 
In practice, non-experts may have to spend substantial effort in order to learn to control an agent.\footnote{Third-person or observational \IL \citep{stadie2017third,sun2019provableoil} allows the teacher to demonstrate tasks with their action space. However, this framework is \textit{non-interactive} because the agent imitates pre-collected demonstrations and does not interact with a teacher. We consider interactive IL \citep{ross2011reduction}, which is shown to be more effective than non-interactive counterparts.} 
In these settings, the agent usually learns from relatively few demonstrations because it does not have to learn to interpret feedback, and the feedback directly specifies the desired behavior. 
At another extreme, we have \RL and \mainsetup, where the teacher provides feedback via \textit{agent-agnostic} media (reward and language, respectively). 
\RL eliminates the agent communication-learning effort by hard-coding the semantics of scalar rewards into the learning algorithm.\footnote{By design, \RL algorithms understand that higher reward value implies better performance.} 
But the trade-off of using such limited feedback is that the task-learning effort of the agent increases; state-of-the-art $\RL$ algorithms are notorious for their high sample complexity ~\cite{Hermann2017,chaplot2017gated,chevalier2018babyai}.
By employing a natural and expressive medium like natural language, \mainsetup offers a compromise between \RL and \IL: it can be more sample-efficient than \RL while not requiring the teacher to master the agent's control interface as \IL does.   
Overall, no protocol is superior in all metrics and the choice of protocol depends on users' preferences.

We empirically evaluate $\mainalg$ against $\IL$ and $\RL$ baselines on two tasks: vision-language navigation \citep{anderson2018vision}, and word-modification via regular expressions \citep{andreas-etal-2018-learning}.
Our results show that $\mainalg$ significantly outperforms $\RL$ baselines in terms of both sample efficiency and quality of the learnt policies.
Also, $\mainalg$'s success rate is competitive with those of the \IL baselines on the navigation task and is lower by 4\% on the word modification task. 
It takes approximately 5-9 times more training episodes than the \IL baselines to reach comparable success rates, which is quite respectable considering that the algorithm has to search in an exponentially large space for the ground-truth executions whereas the \IL baselines are \textit{given} these executions. 
Therefore, $\mainalg$ can be a preferred algorithm whenever annotating executions with correct (agent) actions is not feasible or is substantially more expensive than describing executions in some description language. 
For example, in the word-modification task, $\mainalg$ teaches the agent without requiring a teacher with knowledge about regular expressions.
We believe the capability of non-experts to provide feedback will make $\mainalg$ and more generally the $\mainsetup$ protocol a strong contender in many scenarios. 
The code of our experiments is available at \url{https://github.com/khanhptnk/iliad}.

\section{$\mainsetup$: Interactive Learning from Activity Description}

\paragraph{Environment.} We borrow our terminology from the reinforcement learning (RL) literature~\cite{sutton2018reinforcement}. 
We consider an agent acting in an environment with state space $\Scal$, action space $\Acal$, and transition function $T: \Scal \times \Acal \rightarrow \Delta(\Scal)$, where $\Delta(\Scal)$ denotes the space of all probability distributions over $\Scal$. Let $\Rcal = \left\{R : \Scal \times \Acal \rightarrow [0, 1] \right\}$ be a set of reward functions. 
A \textit{task} in the environment is defined by a tuple $\left( R, s_1, d^{\star} \right)$, where $R \in \Rcal$ is the task's reward function, $s_1 \in \Scal$ is the start state, and $d^{\star} \in \Dcal$ is the task's (language) request. Here, $\Dcal$ is the set of all nonempty strings generated from a finite vocabulary.
The agent only has access to the start state and the task request; the reward function is only used for evaluation.
For example, in robot navigation, a task is given by a start location, a task request like ``\textit{go to the kitchen}", and a reward function that measures the distance from a current location to the kitchen.

\paragraph{Execution Episode.} 
At the beginning of an episode, a task $q = \left( R, s_1, d^{\star}\right)$ is sampled from a task distribution $\PP^{\star}(q)$. The agent starts in $s_1$ and is presented with $d^\star$ but \emph{does not} observe $R$ or any rewards generated by it. The agent maintains a \emph{request-conditioned policy} $\pitheta : \Scal \times \Dcal \rightarrow \Delta(\Acal)$ with parameters $\theta$, which takes in a state $s \in \Scal$ and a request $d \in \Dcal$, and outputs a probability distribution over $\Acal$.
Using this policy, it can generate an \textit{execution} $\hat{e} = \left(s_1, \hat{a}_1, s_2, \cdots, s_H, \hat{a}_H \right)$, where $H$ is the task horizon (the time limit), $\hat{a}_i \sim \pitheta\left( \cdot \mid s_i, d^{\star} \right)$ and $s_{i+1} \sim T\left( \cdot \mid s_i, \hat{a}_i \right)$ for every $i$. Throughout the paper, we will use the notation $e \sim \PP_{\pi}\left(\cdot \mid s_1, d\right)$ to denote sampling an execution $e$ by following policy $\pi$ given a start state $s_1$ and a request $d$. 
The objective of the agent is to find a policy $\pi$ with maximum value, where we define the policy value $V(\pi)$ as:
\begin{align}
    V(\pi) =  \mathbb{E}_{q \sim \PP^{\star}(\cdot), \hat{e} \sim \PP_{\pi}\left(\cdot \mid s_1, d^{\star} \right)} \left[ \sum_{i = 1}^H R\left(s_i, \hat{a}_i\right) \right]
\label{eqn:value_function}
\end{align}

\begin{algorithm}[t!]
\small
\caption{\small $\mainsetup$ protocol. Details of \pref{line:setup-generate-execution} and \pref{line:setup-update} are left to specific implementations.}
\label{alg:protocol}
\begin{algorithmic}[1]
\State Initialize agent policy $\pi_{\theta}: \Scal \times \Dcal \rightarrow \Delta(\Acal)$
\For{$n = 1, 2, \cdots, N$} \label{line:setup-for-loop-start}
\State \!\!World samples a task $q = (R, s_1, d^{\star}) \sim \PP^{\star}(\cdot)$ \label{line:setup-task-given}
\State \!\!\textcolor{redbrown}{Agent generates an execution $\hat{e}$ given $s_1$, $d^\star$, and $\pi_\theta$} \label{line:setup-generate-execution}
\State \!\!Teacher generates a description $\hat{d} \sim \PP_T\left(\cdot \mid \hat{e} \right)$ \label{line:setup-generate-description}
\State \!\!\textcolor{redbrown}{Agent uses $(d^\star, \hat e, \hat{d})$ to update $\pitheta$} \label{line:setup-update}
\EndFor \label{line:setup-for-loop-ends}
\Return $\pitheta$
\end{algorithmic}
\end{algorithm}

\noindent\textbf{$\mainsetup$ protocol.} \autoref{alg:protocol} describes the $\mainsetup$ protocol for training a request-fulfilling agent. It consists of a series of $N$ training episodes.
Each episode starts with sampling a task $q=(R, s_1, d^\star)$ from $\PP^\star$. 
The agent then generates an execution $\hat{e}$ given $s_1$, $d^{\star}$, and its policy $\pi_{\theta}$ (\pref{line:setup-generate-execution}). The feedback mechanism in $\mainsetup$ is provided by a \textit{teacher} that can describe executions in a description language.
The teacher is modeled by a fixed distribution $\PP_T: (\Scal \times \Acal)^H \rightarrow \Delta(\Dcal)$, where $(\Scal \times \Acal)^H$ is the space over $H$-step executions.  
After generating $\hat{e}$, the agent sends it to the teacher and receives a \textit{description} of $\hat{e}$, which is a sample $\hat{d} \sim \PP_T(\cdot \mid \hat{e})$ (\pref{line:setup-generate-description}). 
Finally, the agent uses the triplet $( d^\star, \hat{e}, \hat{d})$ to update its policy for the next round (\pref{line:setup-update}). Crucially, the agent \emph{never} receives any other feedback, including rewards, demonstrations, constraints, or direct knowledge of the \emph{latent} reward function. Any algorithm implementing the \mainsetup protocol has to decide how to generate executions (the exploration problem, \pref{line:setup-generate-execution}) and how to update the agent policy (the grounding problem, \pref{line:setup-update}). 
The protocol does not provide any constraints for these decisions.

\noindent\textbf{Consistency of the teacher.} 
In order for the agent to learn to execute requests by grounding the description language, we
require that the description language is similar to the request language.
Formally, we define the ground-truth joint distribution over tasks and executions as follows
\begin{equation}
    \PP^\star\left(e, R, s_1, d\right) = \PP_{\pi^\star}\left(e \mid s_1, d\right) \PP^\star\left(R, s_1, d\right)\label{eqn:joint-dist}
\end{equation} where $\pi^{\star}$ is an optimal policy that maximizes \autoref{eqn:value_function}.
From this joint distribution, we derive the ground-truth execution-conditioned distribution over requests $\PP^\star(d \mid e)$.
This distribution specifies the probability that a request $d$ can serve as a valid description of an execution $e$.

We expect that if the teacher's distribution $\PP_{T}(d \mid e)$ is close to $\PP^{\star}(d \mid e)$ then grounding the description language to executions will help with request fulfilling.
In that case, the agent can treat a description of an execution as a request that is fulfilled by that execution. 
Therefore, the description-execution pairs $(\hat{d}, \hat{e})$ can be used as supervised-learning examples for the request-fulfilling problem.

The learning process can be sped up if the agent is able to exploit the compositionality of language. For example, if a request is ``\textit{turn right, walk to the kitchen}" and the agent's execution is described as ``\textit{turn right, walk to the bedroom}", the agent may not have successfully fulfilled the task but it can learn what ``\textit{turn right}" and ``\textit{walk to}" mean through the description. Later, it may learn to recognize ``\textit{kitchen}" through a description like ``\textit{go to the kitchen}" and compose that knowledge with its understanding of ``\textit{walk to}" to better execute ``\textit{walk to the kitchen}".

\section{\mainalg: Learning from Activity Describers via Semi-Supervised Exploration}
\label{sec:adel}

We frame the $\mainsetup$ problem as a density-estimation problem: \textit{given that we can effectively draw samples from the distribution $\PP^{\star}(s_1, d)$ and a teacher $\PP_T(d \mid e)$, how do we learn a policy $\pitheta$ such that  $\PP_{\pitheta}(e \mid s_1, d)$ is close to $\PP^{\star}(e \mid s_1, d)$?} 
Here, $\PP^{\star}(e \mid s_1, d) = \PP_{\pi^{\star}}(e \mid s_1, d)$ is the ground-truth request-fulfilling distribution obtained from the joint distribution defined in \autoref{eqn:joint-dist}.

If $s_1$ is not the start state of $e$, then $\PP^{\star}(e \mid s_1, d) = 0$. 
Otherwise, by applying Bayes' rule, and noting that $s_1$ is included in $e$, we have:
\begin{align}
\notag
    \PP^{\star}(e \mid s_1, d)
    &
    \propto \PP^{\star}(e, d \mid s_1)
    = \PP^{\star}(e \mid s_1) \PP^{\star}(d \mid e, s_1),
\\
\notag
    &
    = \PP^{\star}(e \mid s_1) \PP^{\star}(d \mid e),
\\
    &
    \approx \PP^{\star}(e \mid s_1)\PP_T(d \mid e).
\end{align}
As seen from the equation, the only missing piece required for estimating $\PP^{\star}(e \mid s_1, d)$ is the marginal\footnote{
We are largely concerned with the relationship between $e$ and $d$,
and so refer to the distribution $\PP^{\star}(e \mid s_1)$ as the marginal and $\PP^{\star}(e \mid s_1, d)$ as the conditional.
}
$\PP^{\star}(e \mid s_1 )$.
\autoref{alg:true_marginal} presents a simple method for learning an agent policy if we have access to this marginal.
It is easy to show that the pairs $(\hat{e}, \hat{d})$ in the algorithm are approximately drawn from the joint distribution $\PP^{\star}(e, d \mid s_1)$ and thus can be directly used to estimate the conditional $\PP^{\star}(e \mid s_1, d)$.

\begin{algorithm}[t!]
\small
\caption{\small Simple algorithm for learning an agent's policy with access to the true marginal $\PP^{\star}(e \mid s_1)$ and teacher $\PP_T(d \mid e)$.}
\label{alg:true_marginal}
\begin{algorithmic}[1]
\State $\Bcal = \emptyset$ 
\For{$i = 1, 2, \cdots, N$}
\State World samples a task $q = (R, s_1, d^{\star}) \sim \PP^{\star}(\cdot)$
\State Sample $(\hat{e}, \hat{d})$ as follows: $\hat{e} \sim \PP^{\star}(\cdot \mid s_1), \hat{d} \sim \PP_T(\cdot \mid \hat{e})$
\State $\Bcal \leftarrow \Bcal \cup \{ ( \hat{e}, \hat{d}) \}$
\EndFor
\State  Train a policy $\pitheta(a \mid s, d)$ via maximum log-likelihood: \par $\max_{\theta} \sum_{\left(\hat{e}, \hat{d}\right) \in \Bcal} \sum_{(s, \hat a_s) \in \hat{e}} \log \pitheta (\hat a_s \mid s, \hat{d})$ \par where $\hat a_s$ is the action taken by the agent in state $s$
\State \Return $\pitheta$ 
\end{algorithmic}
\end{algorithm}

Unfortunately, $\PP^{\star}(e \mid s_1)$ is unknown in our setting. 
We present our main algorithm \mainalg (\autoref{alg:adel}) which simultaneously estimates $\PP^{\star}(e \mid s_1)$ and $\PP^{\star}(e \mid s_1, d)$ through interactions with the teacher.
In this algorithm, we assume access to an \textit{approximate marginal} $\PP_{\piomega}(e \mid s_1)$ defined by an \textit{explorative policy} $\piomega\left(a \mid s\right)$. 
This policy can be learned from a dataset of unlabeled executions or be defined as a program that synthesizes executions. 
In many applications, reasonable unlabeled executions can be cheaply constructed using knowledge about the structure of the execution. 
For example, in robot navigation, valid executions are collision-free and non-looping; in semantic parsing, predicted parses should follow the syntax of the semantic language.

After constructing the approximate marginal $\PP_{\piomega}(e \mid s_1)$, we could substitute it for the true marginal in \autoref{alg:true_marginal}. However, using a fixed approximation of the marginal may lead to sample inefficiency when there is a mismatch between the approximate marginal and the true marginal. 
For example, in the robot navigation example, if most human requests specify the kitchen as the destination, the agent should focus on generating executions that end in the kitchen to obtain descriptions that are similar to those requests.
If instead, a uniform approximate marginal is used to generate executions, the agent obtains a lot of irrelevant descriptions.

\begin{algorithm}[t!]
\small
\caption{\small $\mainalg$: our implementation of the $\mainsetup$ protocol.}
\label{alg:adel}
\begin{algorithmic}[1]
\State \textbf{Input}: teacher $\PP_T(d \mathbin{\mid} e)$, approximate marginal $\PP_{\piomega}(e \mathbin{\mid} s_1)$, mixing weight $\lambda \in [0, 1]$, annealing rate $\beta \in (0, 1)$
\State Initialize $\pi_{\theta}: \Scal \times \Dcal \rightarrow \Delta(\Acal)$ and $\Bcal = \emptyset$ \label{line:mainalg-policy-init}
\For{$n = 1, 2, \cdots, N$} \label{line:mainalg-for-loop-starts}
\State World samples a task $q=(R, s_1, d^{\star}) \sim \PP^{\star}(\cdot)$ \label{line:mainalg-sample-task}
\State Agent generates $\hat{e} \sim \tilde{\PP}(\cdot \mid s_1, d^{\star})$ (see \autoref{eqn:mix_marginal})   \label{line:mainalg-sample-execution}
\State Teacher generates a description $\hat{d} \sim \PP_T(\cdot \mid \hat{e})$  \label{line:mainalg-sample-description}
\State $\Bcal \leftarrow \Bcal \cup (\hat{e}, \hat{d})$ \label{line:mainalg-add-datapoint}
\State Update agent policy:
\begin{align}
    \theta \leftarrow \max_{\theta'} \sum_{(\hat{e}, \hat{d}) \in \Bcal} \sum_{(s, \hat a_s) \in \hat{e}} \log \pi_{\theta'} (\hat a_s \mid s, \hat{d} ) \nonumber
\end{align} \par where $\hat a_s$ is the action taken by the agent in state $s$ \label{line:mainalg-update}
\State Anneal mixing weight: $\lambda \leftarrow \lambda\cdot\beta $
\EndFor \label{line:mainalg-for-loop-ends}
\State \Return $\pitheta$
\end{algorithmic}
\end{algorithm}

\mainalg minimizes potential marginal mismatch
by iteratively using the estimate of the marginal $\PP^{\star}(e \mid s_1)$ to improve the estimate of the conditional $\PP^{\star}(e \mid s_1, d)$ and vice versa.
Initially, we set $\PP_{\piomega}(e \mid s_1)$ as the marginal over executions.
In each episode, we \emph{mix} this distribution with $\PP_{\pitheta}(e \mid s_1, d)$, the current estimate of the conditional, to obtain an improved estimate of the marginal (\pref{line:mainalg-sample-execution}).
Formally, given a start state $s_1$ and a request $d^{\star}$, we sample an execution $\hat{e}$ from the following distribution: 
\begin{align}
    \tilde{\PP}(\cdot \mid s_1, d^{\star}) \triangleq   \lambda\PP_{\piomega}(\cdot \mid s_1) + 
    (1  - \lambda)\PP_{\pitheta}(\cdot \mid s_1, d^{\star})
\label{eqn:mix_marginal}
\end{align}
where $\lambda \in [0, 1]$ is a mixing weight that is annealed to zero over the course of training.
Each component of the mixture in \autoref{eqn:mix_marginal} is essential in different learning stages. 
Mixing with $\PP_{\piomega}$ accelerates convergence at the early stage of learning.
Later, when $\pitheta$ improves, $\PP_{\pitheta}$ skews $\tilde{\PP}$ 
towards executions whose descriptions are closer to the requests, closing the gap with $\PP^{\star}(e \mid s_1)$.
In \pref{line:mainalg-sample-description}-\ref{line:mainalg-update}, similar to \autoref{alg:true_marginal}, we leverage the (improved) marginal estimate and the teacher to draw samples $( \hat{e}, \hat{d})$ and use them to re-estimate $\PP_{\pitheta}$.

\paragraph{Theoretical Analysis.}
We analyze an epoch-based variant of $\mainalg$ and show that under certain assumptions, it converges to a near-optimal policy. 
In this variant, we 
run the algorithm in epochs, where the agent policy is only updated at the end of an epoch. In each epoch, we collect a fresh batch of examples $\{( \hat{e}, \hat{d})\}$ as in $\mainalg$ (\pref{line:mainalg-sample-task}-\ref{line:mainalg-add-datapoint}), and use them to perform a batch update (\pref{line:mainalg-update}).
We provide a sketch of our theoretical results here and defer the full details to~\pref{app:theory-analysis}. 

We consider the case of $H=1$ where an execution $e=(s_1, a)$ consists of the start state $s_1$ and a single action $a$ taken by the agent. This setting while restrictive captures the non-trivial class of contextual bandit problems~\citep{langford2008cb}. 
Sequential decision-making problems where the agent makes decisions solely based on the start state can be reduced to this setting by treating a sequence of decisions as a single action~\citep{kreutzer-etal-2017-bandit,nguyen-etal-2017-reinforcement}. We focus on the convergence of the iterations of epochs, and assume that the maximum likelihood estimation problem in each epoch can be solved optimally. We also ablate the teacher learning difficulty by assuming access to a perfectly consistent teacher, i.e., $\PP_T(d \mid e) = \PP^\star(d \mid e)$. 

We make two crucial assumptions. Firstly, we make a standard realizability assumption to ensure that our policy class is expressive enough to accommodate the optimal solution of the maximum likelihood estimation. Secondly, we assume that for every start state $s_1$, the teacher distribution's matrix $\PP^\star(d \mid e_{s_1})$ over  descriptions and executions $e_{s_1}$ starting with $s_1$, has a non-zero minimum singular value $\smin(s_1)$. Intuitively, this assumption implies that descriptions are rich enough to help in deciphering actions. 
Under these assumptions, we prove the following result:

\begin{theorem}[Main Result]\label{thm:main-theorem} Let $\PP_n(e \mid s_1)$ be the marginal distribution in the $n^{th}$ epoch. Then for any $t \in \NN$ and any start state $s_1$ we have:
\begin{equation*}
     \|\PP^\star(e \mid s_1) - \frac{1}{t} \sum_{n=1}^{t} \PP_n(e \mid s_1)\|_2 \le \frac{1}{\smin(s_1)}\sqrt{\frac{2\ln |\Acal|}{ t}}.
\end{equation*}
\end{theorem}
\pref{thm:main-theorem} shows that the running average of the estimated marginal distribution converges to the true marginal distribution. The error bound depends logarithmically on the size of action space, and therefore, suitable for problems with exponentially large action space. As argued before, access to the true marginal can be used to easily learn a near-optimal policy. For brevity, we defer the proof and other details to~\pref{app:theory-analysis}. 
Hence, our results show that under certain conditions, we can expect convergence to the optimal policy. 
We leave the question of sample complexity and addressing more general settings for future work.

\section{Experimental Setup}

In this section, we present a general method for simulating an execution-describing teacher using a pre-collected dataset (\autoref{sec:simulation}).
Then we describe setups of the two problems we conduct experiments on: vision-language navigation (\autoref{sec:nav}) and word modification (\autoref{sec:regex}). 
Details about the data, the model architecture, training hyperparameters, and how the teacher is simulated in each problem are in the Appendix.

We emphasize that the \mainsetup protocol or the \mainalg algorithm do \textit{not} propose learning a teacher. 
Similar to \IL and \RL, \mainsetup operates with a fixed, black-box teacher that is given in the environment. 
Our experiments specifically simulate \textit{human} teachers that train request-fulfilling agents by talking to them (using descriptions). 
We use labeled executions only to learn approximate models of human teachers.\looseness=-1

\subsection{Simulating Teachers}
\label{sec:simulation}

$\mainsetup$ assumes access to a teacher $\PP_T(d \mid e)$ that can describe agent executions in a description language. 
For our experimental purposes, employing human teachers is expensive and irreproducible, thus we simulate them using pre-collected datasets.
We assume availability of a dataset $\Bcal_{\textrm{sim}} = \left\{\left( \Dcal_n^{\star}, e_n^{\star} \right)\right\}_{n = 1}^N$, where $\Dcal_n^{\star} = \{ d_n^{\star(j)} \}_{j = 1}^M$ contains $M$ human-generated requests that are fulfilled by execution $e_n^{\star}$.
Each of the two experimented problems is accompanied by data that is partitioned into training/validation/test splits. 
We use the training split as $\Bcal_{\textrm{sim}}$ and use the other two splits for validation and testing, respectively.
Our agents do \textit{not} have direct access to $\Bcal_{\textrm{sim}}$.
From an agent's perspective, it communicates with a black-box teacher that can return descriptions of its executions; it does not know how the teacher is implemented.

Each \mainsetup episode (\autoref{alg:protocol}) requires providing a request $d^{\star}$ at the beginning and a description $\hat{d}$ of an execution $\hat{e}$. 
The request $d^{\star}$ is chosen by first uniformly randomly selecting an example $\left( \Dcal_{n}^{\star}, e_{n}^{\star}\right)$ from $\Bcal_{\textrm{sim}}$, and then uniformly sampling a request $d^{\star(j)}_n$ from $\Dcal_n^{\star}$.
The description $\hat{d}$ is generated as follows.
We first gather all the pairs $(d_n^{\star(j)}, e_n^{\star})$ from $\Bcal_{\textrm{sim}}$ and train an RNN-based conditional language model $\tilde{\PP}_T(d \mid e)$ via standard maximum log-likelihood. 
We can then generate a description of an execution $\hat e$ by greedily decoding\footnote{Greedily decoding an RNN-based model refers to stepwise choosing the highest-probability class of the output softmax. In this case, the classes are words in the description vocabulary.} this model conditioned on $\hat e$: $\hat{d}_{\textrm{greedy}} = \texttt{greedy}\bigl( \tilde{\PP}_T(\cdot \mid \hat{e})\bigr)$.
However, given limited training data, this model may not generate sufficiently high-quality descriptions. 
We apply two techniques to improve the quality of the descriptions. 
First, we provide the agent with the human-generated requests in $\Bcal_{\textrm{sim}}$ when the executions are near optimal.
Let $\texttt{perf}\left(\hat{e}, e^{\star}\right)$\footnote{The metric \texttt{perf} is only used in simulating the teachers and is not necessarily the same as the reward function $R$.} be a performance metric that evaluates an agent's execution $\hat{e}$ against a ground-truth $e^{\star}$ (higher is better).
An execution $\hat e$ is near optimal if $\texttt{perf}\left(\hat{e}, e^{\star}\right) \geq \tau$, where $\tau$ is a constant threshold. 
Second, we apply pragmatic inference \citep{andreas-klein-2016-reasoning,fried-etal-2018-unified}, leveraging the fact that the teacher has access to the environment's simulator and can simulate executions of descriptions.
The final description given to the agent is
\begin{align}
    \hat{d} \sim \begin{cases}
    \texttt{Unif}\left( \Dcal_n^{\star} \right)  &\text{if $\texttt{perf}\left(\hat{e}, e_n^{\star}\right) \geq \tau$,} \\
    \texttt{Unif}\left(\Dcal_{\textrm{prag}} \cup \{ \emptyset \}\right) &\text{otherwise}
     \end{cases}
\end{align} where $\texttt{Unif}(\Dcal)$ is a uniform distribution over elements of $\Dcal$, $e_n^{\star}$ is the ground-truth execution associated with $\Dcal_n^{\star}$, $\Dcal_{\textrm{prag}}$ contains descriptions generated using pragmatic inference (which we will describe next), and $\emptyset$ is the empty string.

\paragraph{Improved Descriptions with Pragmatic Inference.} 
Pragmatic inference emulates the teacher's ability to mentally simulate task execution. 
Suppose the teacher has its own execution policy $\pi_T(a \mid s, d)$, which is learned using the pairs $\left(e_n^{\star}, d_n^{\star(j)}\right)$ of $\Bcal_{\textrm{sim}}$, and access to a simulator of the environment. 
A \textit{pragmatic} execution-describing teacher is defined as $\PP_T^{\textrm{prag}}(d \mid e) \propto \PP_{\pi_T}(e \mid s_1, d)$.
For this teacher, the more likely that a request $d$ 
causes it to generate an execution $e$, the more likely that it describes $e$ as $d$.

In our problems, constructing the pragmatic teacher's distribution explicitly is not feasible because we would have to compute a normalizing constant that sums over all possible descriptions.
Instead, we follow \citet{andreas-etal-2018-learning}, generating a set of candidate descriptions and using $\PP_{\pi_T}(e \mid s_1, d)$ to re-rank those candidates.
Concretely, for every execution $\hat{e}$ where $\texttt{perf}\left(\hat{e}, e_n^{\star}\right) < \tau$, we use the learned language model $\tilde{\PP}_T$ to generate a set of candidate descriptions $\Dcal_{\textrm{cand}} = \{ \hat{d}_{\textrm{greedy}} \} \cup \{ \hat{d}_{\textrm{sample}}^{(k)} \}_{k = 1}^K$.
This set consists of the greedily decoded description $\hat{d}_{\textrm{greedy}} = \texttt{greedy}\left( \tilde{\PP}_T(\cdot \mid \hat{e})\right)$ and $K$ descriptions $\hat{d}_{\textrm{sample}}^{(k)} \sim \tilde{\PP}_T(\cdot \mid \hat{e})$. 
To construct $\Dcal_{\textrm{prag}}$, we select descriptions in $\Dcal_{\textrm{cand}}$ from which $\pi_T$ generates executions that are similar enough to $\hat e$:
\begin{align}
    \Dcal_{\textrm{prag}} = \left\{ d \mid d \in \Dcal_{\textrm{cand}} \wedge  \texttt{perf}\left(e^d, \hat{e} \right) \geq \tau \right\} 
\label{eqn:pragmatic}
\end{align} where $e^d = \texttt{greedy}\left( \PP_{\pi_T}(\cdot \mid s_1, d)\right)$ and $s_1$ is the start state of $\hat{e}$.  

\subsection{Vision-Language Navigation (\nav)}
\label{sec:nav}

\paragraph{Problem and Environment.} An agent executes natural language requests (given in English) by navigating to locations in environments that photo-realistically emulate residential buildings \citep{anderson2018vision}. 
The agent successfully fulfills a request if its final location is within three meters of the intended goal location. 
Navigation in an environment is framed as traversing in a graph where each node represents a location and each edge connects two nearby unobstructed locations.
A state $s$ of an agent represents its location and the direction it is facing. 
In the beginning, the agent starts in state $s_1$ and receives a navigation request $d^{\star}$. 
At every time step, the agent is not given the true state $s$ but only receives an observation $o$, which is a real-world RGB image capturing the panoramic view at its current location. 

\paragraph{Agent Policy.} 
The agent maintains a policy $\pi_{\theta}\left( a \mid o, d \right)$ that takes in a current observation $o$ and a request $d$, and outputs an action $a \in V_{\textrm{adj}}$, where $V_{\textrm{adj}}$ denotes the set of locations that are adjacent to the agent's current location according to the environment graph. 
A special \texttt{<stop>} action is taken when the agent wants to terminate an episode or when it has taken $H$ actions.

\noindent\textbf{Simulated Teacher}. We simulate a teacher that does not know how to control the navigation agent and thus cannot provide demonstrations. 
However, the teacher can verbally describe navigation paths taken by the agent. 
We follow \autoref{sec:simulation}, constructing a teacher $\PP_T(d \mid e)$ that outputs language descriptions given executions $e = \left(o_1, a_1, \cdots, o_H \right)$.

\subsection{Word Modification (\regex)} 
\label{sec:regex}

\paragraph{Problem.} A human gives an agent a natural language request (in English) $d^{\star}$ asking it to modify the characters of a word $w^{\textrm{inp}}$.
The agent must execute the request and outputs a word $\hat w^{\textrm{out}}$.
It successfully fulfills the request if $\hat{w}^{\textrm{out}}$ exactly matches the expected output $w^{\textrm{out}}$.
For example, given an input word \textit{embolden} and a request ``\textit{replace all n with c}", the expected output word is \textit{embolde\underline{c}}.
We train an agent that solves this problem via a semantic parsing approach.
Given $w^{\textrm{inp}}$ and $d^{\star}$, the agent generates a regular expression $\hat{a}_{1:H} = \left( \hat{a}_1, \cdots, \hat{a}_H \right)$, which is a sequence of characters. It then uses a regular expression compiler to apply the regular expression onto the input word to produce an output word $\hat{w}^{\textrm{out}} = \texttt{compile}\left( w^{\textrm{inp}}, \hat{a}_{1:H} \right)$.

\paragraph{Agent Policy and Environment.} 
The agent maintains a policy $\pi_{\theta}\left( a \mid s, d \right)$ that takes in a state $s$ and a request $d$, and outputs a distribution over characters $a \in V_{\textrm{regex}}$, where $V_{\textrm{regex}}$ is the regular expression (character) vocabulary. 
A special \texttt{<stop>} action is taken when the agent wants to stop generating the regular expression or when the regular expression exceeds the length limit $H$. 
We set the initial state $s_1 = \left(w^{\textrm{inp}}, \emptyset \right)$, where $\emptyset$ is the empty string.
A next state is determined as follows
\begin{align}
    s_{t + 1} = \begin{cases}
    \left(\hat{w}^{\textrm{out}}, \hat{a}_{1:t}\right) &\text{if } \hat{a}_t = \texttt{<stop>}, \\
    \left(w^{\textrm{inp}}, \hat{a}_{1:t} \right) &\text{otherwise}
    \end{cases}
\end{align} where $\hat{w}^{\textrm{out}} = \texttt{compile}\left( w^{\textrm{inp}}, \hat{a}_{1:t} \right)$.

\paragraph{Simulated Teacher.} We simulate a teacher that does not have knowledge about regular expressions.
Hence, instead of receiving full executions, which include regular expressions $\hat a_{1:H}$ predicted by the agent, the teacher generates descriptions given only pairs $(w^{\textrm{inp}}_j, \hat{w}^{\textrm{out}}_j)$ of an input word $w^{\textrm{inp}}_j$ and the corresponding output generated by the agent $\hat{w}^{\textrm{out}}_j$.
In addition, to reduce ambiguity, the teacher requires multiple word pairs generate a description. 
This issue is better illustrated in the following example. 
Suppose the agent generates the pair \textit{embolden} $\rightarrow$ \textit{emboldec} by predicting a regular expression that corresponds to the description ``\textit{replace all n with c}". However, because the teacher does not observe the agent's regular expression (and cannot understand the expression even if it does), it can also describe the pair as ``\textit{replace the last letter with c}". 
Giving such a description to the agent would be problematic because the description does not correspond to the predicted regular expression. 
Observing multiple word pairs increases the chance that the teacher's description matches the agent's regular expression (e.g. adding \textit{now} $\rightarrow$ \textit{cow} help clarify that ``\textit{replace all n with c}" should be generated).
In the end, the teacher is a model $P_T \bigl(d \bigm| \{ w^{\textrm{inp}}_j, \hat{w}^{\textrm{out}}_j \}_{j = 1}^J \bigr)$
that takes as input $J$ word pairs.
To generate $J$ word pairs, in every episode, in addition to the episode's input word, we sample $J - 1$ more words from the dictionary and execute the episode's request on the $J$ words.
We do \textit{not} use any regular expression data in constructing the teacher.
To train the teacher's policy $\pi_T$ for pragmatic inference (\autoref{sec:simulation}), we use a dataset that consists of tuples $\left(\Dcal_n^{\star}, (w^{\textrm{inp}}_n, w^{\textrm{out}}_n ) \right)$ which are not annotated with ground-truth regular expressions. 
$\pi_T$ directly generates an output word instead of predicting a regular expression like the agent policy $\pi_{\theta}$.

\begin{figure}[t!]
    \centering
    \begin{subfigure}[t]{\linewidth}
        \includegraphics[width=0.95\linewidth]{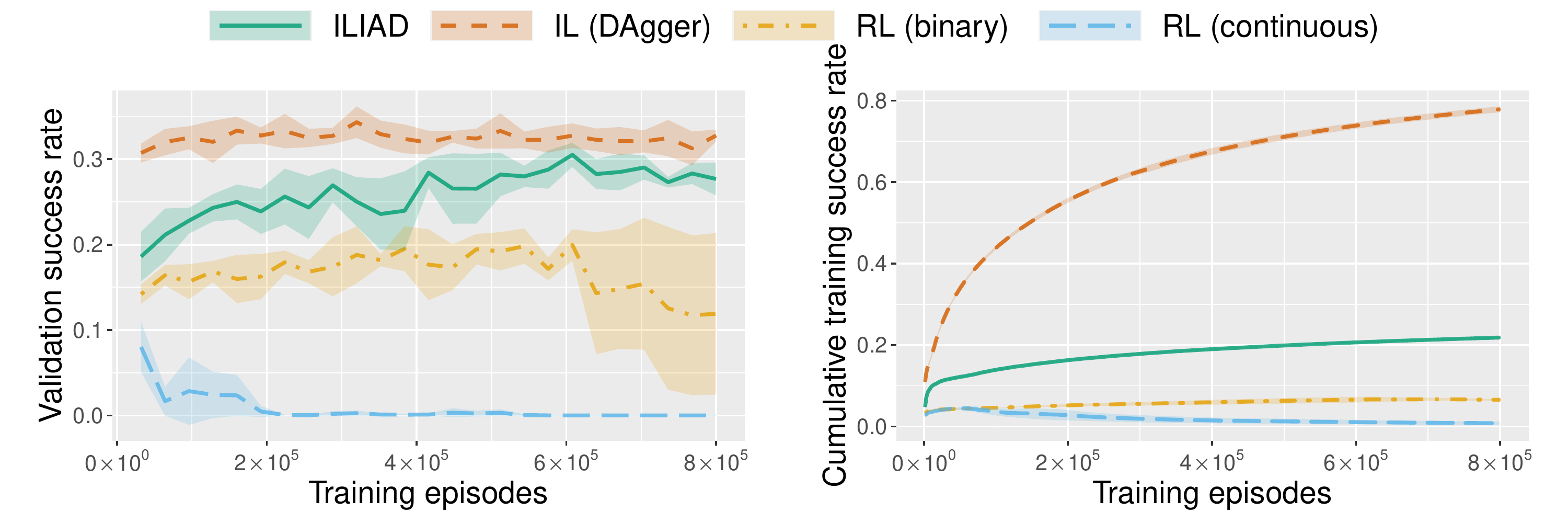}
        \caption{\nav}
    \end{subfigure}
    ~
    \begin{subfigure}[t]{\linewidth}
        \includegraphics[width=0.95\linewidth]{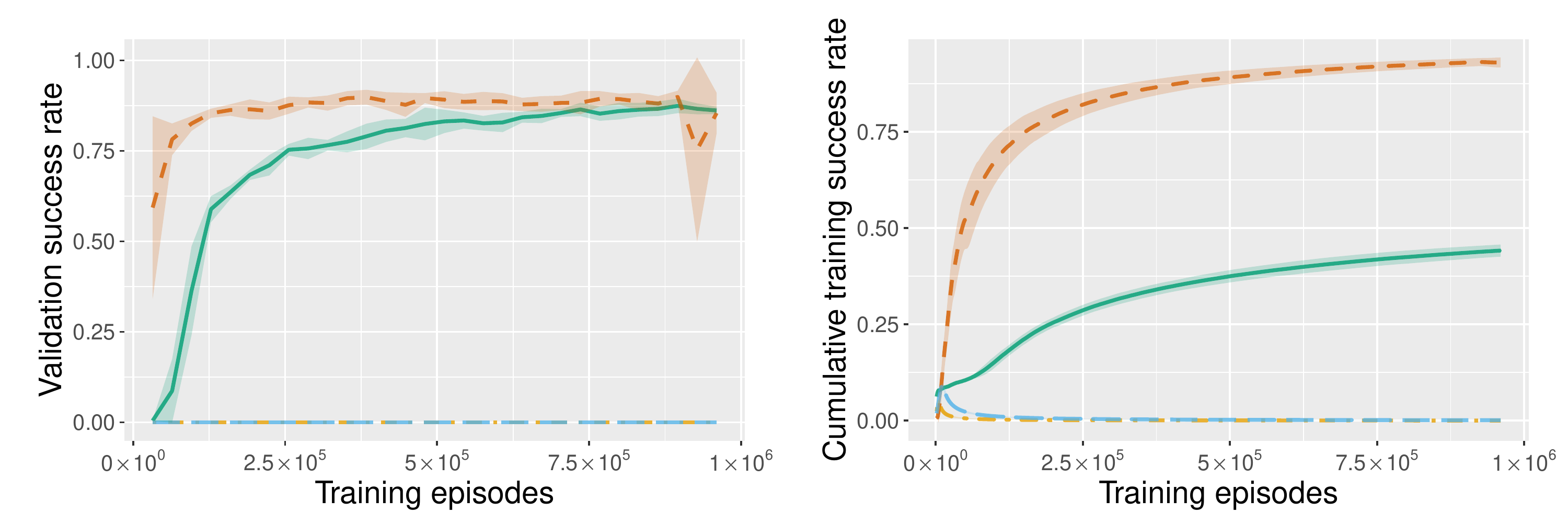}
        \caption{\regex}
    \end{subfigure}
    \caption{Validation success rate (average held-out return) and cumulative training success rate (average training return) over the course of training. For each algorithm, we report means and standard deviations over five runs with different random seeds.}
    \label{fig:main}
\end{figure}

\subsection{Baselines and Evaluation Metrics}

We compare interactive learning settings that employ different teaching media:
\begin{itemize}
    \item \textit{Learning from activity description} ($\mainsetup$): the teacher returns a language description $\hat{d}$. 
    \item \textit{Imitation learning} (\IL): 
    the teacher demonstrates the correct actions in the states that the agent visited, returning $e^{\star} = \left( s_1, a^{\star}_1, \cdots, a^{\star}_H, s_H \right)$, where $s_i$ are the states in the agent's execution and $a^{\star}_i$ are the optimal actions in those states.
    \item \textit{Reinforcement learning} (\RL): the teacher provides a scalar reward that evaluates the agent's execution. We consider a special case when rewards are provided only at the end of an episode. 
    Because such feedback is cheap to collect (e.g., star ratings) \citep{nguyen2017banditnmt,kreutzer-etal-2018-reliability,kreutzer2020learning}, this setting is suitable for large-scale applications. 
    We experiment with both binary reward that indicates task success, and continuous reward that measures  normalized distance to the goal (see \autoref{app:training}). 
\end{itemize} 
We use $\mainalg$ in the $\mainsetup$ setting,  DAgger \citep{ross2011reduction} in \IL, and \reinforce\footnote{We use a moving-average baseline to reduce variance. We also experimented with A2C \citep{mnih2016asynchronous} but it was less stable in this sparse-reward setting. At the time this paper was written, we were not aware of any work that successfully trained agents using RL without supervised-learning bootstrapping in the two problems we experimented on.} \citep{Williams:92reinforce} in \RL.
We report the \textit{success rates} of these algorithms, which are the fractions of held-out (validation or test) examples on which the agent successfully fulfills its requests. 
All agents are initialized with random parameters.

\begin{table*}[t!]
\centering
\scriptsize
\caption{Main results. We report means and standard deviations of success rates (\%) over five runs with different random seeds. \RL-Binary and \RL-Cont refer to the \RL settings with binary and continuous rewards, respectively. Sample complexity is the number of training episodes (or number of teacher responses) required to reach a validation success rate of at least $c$. 
Note that the teaching efforts are not comparable across the learning settings: providing a demonstration can be more or less tedious than providing a language description depending on various characteristics of the teacher. Hence, even though \mainalg requires more  episodes to reach the same performance as DAgger, we do not draw any conclusions about the primacy of one algorithm over the other in terms of teaching effort. }
\vspace{0.1cm}
\begin{tabular}{ccccccc}
    \toprule
      &  &  &  & \multicolumn{3}{c}{Sample complexity $\downarrow$}   \\ \cmidrule(lr){5-7} 
     Learning setting & Algorithm & Val success rate (\%) $\uparrow$ & Test success rate (\%) $\uparrow$ & \# Demonstrations & \# Rewards & \# Descriptions    \\  \midrule  
     \multicolumn{4}{l}{\textbf{Vision-language navigation}} & \multicolumn{3}{c}{($c$ = 30.0\%)} \\ 
     \IL & DAgger & 35.6 $\pm$ 1.35 & 32.0 $\pm$ 1.63 & ~~45K $\pm$ 26K & - & - \\ 
     \RL-Binary & \textsc{Reinforce} & 22.4 $\pm$ 1.15 & 20.5 $\pm$ 0.58 & - & $+\infty$ & - \\ 
     \RL-Cont & \textsc{Reinforce}  & 11.1 $\pm$ 2.19 & 11.3 $\pm$ 1.25 & - & $+\infty$ & - \\ 
     $\mainsetup$ & $\mainalg$  & 32.2 $\pm$ 0.97 & 31.9 $\pm$ 0.76 & - & - & 406K $\pm$ 31K \\ 
     \midrule
     \multicolumn{4}{l}{\textbf{Word modification}} & \multicolumn{3}{c}{($c$ = 85.0\%)} \\ 
     \IL & DAgger & 92.5 $\pm$ 0.53 & 93.0 $\pm$ 0.37 & 118K $\pm$ 16K & - & -  \\ 
     \RL-Binary & \textsc{Reinforce} & ~~0.0 $\pm$ 0.00 & ~~0.0 $\pm$ 0.00 & - & $+\infty$ & - \\ 
     \RL-Cont & \textsc{Reinforce} & ~~0.0 $\pm$ 0.00 & ~~0.0 $\pm$ 0.00 & - &  $+\infty$ & - \\ 
     $\mainsetup$ & $\mainalg$ & 88.1 $\pm$ 1.60 & 89.0 $\pm$ 1.30 & - & - & 573K $\pm$ 116K \\ 
     \bottomrule
\end{tabular}
\label{tab:main}
\end{table*}

\section{Results}

We compare the learning algorithms on not only success rate, but also the effort expended by the teacher.
While task success rate is straightforward to compute, teacher effort is hard to quantify because it depends on many factors: the type of knowledge required to teach a task, the cognitive and physical ability of a teacher, etc. 
For example, in \regex, providing demonstrations in forms of regular expressions may be easy for a computer science student, but could be challenging for someone who is unfamiliar with programming.  
In \nav, controlling a robot may not be viable for an individual with motor impairment, whereas generating language descriptions may infeasible for someone with a verbal-communication disorder. 
Because it is not possible cover all teacher demographics, our goal is to \textit{quantitatively} compare the learning algorithms on learning effectiveness and efficiency, and  \textit{qualitatively} compare them on the teacher effort to learn to express feedback using the protocol's communication medium.
Our overall findings (\autoref{tab:tradeoff}) highlight the strengths and weaknesses of each learning algorithm and can potentially aid practitioners in selecting algorithms that best suit their applications.

\paragraph{Main results.} Our main results are in \autoref{tab:main}. 
Overall, results in both problems match our expectations. 
The \IL baseline achieves the highest success rates (on average, 35.6\% on \nav and 92.5\% on \regex). 
This framework is most effective because the feedback directly specifies ground-truth actions. 
The RL baseline is unable to reach competitive success rates.
Especially, in \regex, the RL agent cannot learn the syntax of the regular expressions and completely fails at test time.
This shows that the reward feedback is not sufficiently informative to guide the agent to explore efficiently in this problem. 
$\mainalg$'s success rates are slightly lower than those of \IL (3-4\% lower than) but are substantially higher than those of RL (+9.8\% on \nav and +88.1\% on \regex compared to the best RL results).

To measure learning efficiency, we report the number of training episodes required to reach a substantially high success rate (30\% for \nav and 85\% for \regex). 
We observe that all algorithms require hundreds of thousands of episodes to attain those success rates. 
The RL agents cannot learn effectively even after collecting more than 1M responses from the teachers. 
$\mainalg$ attains reasonable success rates using 5-9 times more responses than \IL.
This is a decent efficiency considering that $\mainalg$ needs to find the ground-truth executions in exponentially large search spaces, while \IL directly communicates these executions to the agents. 
As $\mainalg$ lacks access to ground-truth executions, its average training returns are 2-4 times lower than those of \IL  (\autoref{fig:main}). \looseness=-1

\begin{table}[t!]
\centering
\footnotesize
\setlength{\tabcolsep}{1.2pt}
\caption{Effects of mixing execution policies in ADEL.}
\vspace{0.1cm}
\begin{tabular}{lcc}
    \toprule
     \multicolumn{1}{c}{Mixing weight} & Val success rate (\%) $\uparrow$ & Sample complexity $\downarrow$   \\  \midrule
     \multicolumn{3}{l}{\textbf{Vision-language navigation}}
     \\ 
     $\lambda = 0$ (no marginal) & ~0.0 & $+\infty$ \\
     $\lambda = 1$ & 29.4 & $+\infty$ \\
     $\lambda = 0.5$ (final) & 32.0 & 384K \\
     \midrule
     \multicolumn{3}{l}{\textbf{Word modification}} \\
     $\lambda = 0$ (no marginal) & ~0.2 & $+\infty$ \\
     $\lambda = 1$ & 55.7 & $+\infty$ \\
     $\lambda = 0.5$ (final) & 88.0 & 608K \\
     \bottomrule
\end{tabular}
\label{tab:ablation}
\end{table}

\paragraph{Ablation.} 
We study the effects of mixing with the approximate marginal ($\PP_{\piomega}$) in $\mainalg$ (\autoref{tab:ablation}).
First of all, we observe that learning cannot take off without using the approximate marginal ($\lambda$ = 0).
On the other hand, using only the approximate marginal to generate executions ($\lambda = 1$) degrades performance, in terms of both success rate and sample efficiency. 
This effect is more visible on \regex where the success rate drops by 33\% (compared to a 3\% drop in \nav), indicating that the gap between the approximate marginal and the true marginal is larger in \regex than in \nav. 
This matches our expectation as the set of unlabeled executions that we generate to learn $\pi_{\omega}$ in \regex covers a smaller portion of the problem's execution space  than that in \nav. 
Finally, mixing the approximate marginal and the agent-estimated conditional ($\lambda = 0.5$) gives the best results. \looseness=-1

\section{Related Work}

\paragraph{Learning from Language Feedback.} Frameworks for learning from language-based communication have been previously proposed. 
Common approaches include: reduction to reinforcement learning \citep{goldwasser2014learning,macglashan2015grounding, ling2017teaching,goyal2019using,fu2019language,sumers2020learning}, learning to ground language to actions \citep{chen2011learning,misra2014tell,bisk-etal-2016-natural,liu2016jointly,wang-etal-2016-learning-language,li2017programming,li2020towards,li2020interactive}, or devising EM-based algorithms to parse language into logical forms  \citep{matuszek2012grounded,labutov-etal-2018-learning}.
The first approach may discard useful learning signals from language feedback and inherits the limitations of \RL algorithms. 
The second requires extra effort from the teacher to provide demonstrations.
The third approach has to bootstrap the language parser with labeled executions. 
\mainalg enables learning from a specific type of language feedback (language description) without reducing it to reward, requiring demonstrations, or assuming access to labeled executions. 

\paragraph{Description Feedback in Reinforcement Learning.} Recently, several papers have proposed using language description feedback in the context of reinforcement learning \citep{jiang2019language, chan2019actrce, colas2020language, cideron2020higher}.
These frameworks can be viewed as extensions of hindsight experience replay (HER; \citealp{andrychowicz2017hindsight}) to language goal generation. 
While the teacher in \mainsetup can be considered as a language goal generator, an important distinction between \mainsetup and these frameworks is that \mainsetup models a completely \textit{reward-free} setting.
Unlike in HER, the agent in \mainsetup does not have access to a reward function that it can use to compute the reward of any tuple of state, action, and goal. 
With the feedback coming solely from language descriptions, \mainsetup is designed so that task learning relies only on extracting information from language. 
Moreover, unlike reward, the description language in \mainsetup does not contain information that \emph{explicitly} encourages or discourages actions of the agent. 
The formalism and theoretical studies of \mainsetup presented in this work are based on a probabilistic formalism and do not involve reward maximization. 

\paragraph{Description Feedback in Vision-Language Navigation.} Several papers \citep{fried2018speaker,tan-etal-2019-learning} apply back-translation to vision-language navigation \citep{anderson2018vision}.
While also operating with an output-to-input translator, back-translation is a single-round, offline process, whereas \mainsetup is an iterative, online process.  
\citet{Zhou2021InverseRL} study a test-time scenario that is similar to \mainsetup but requires labeled demonstrations to learn the execution describer and to initialize the agent. The teacher in \mainsetup is more general: it can be automated (i.e., learned from labeled data), but it can also be a human. 
Our experiments emulate applications where non-expert humans teach agents new tasks by only giving them verbal feedback.  
We use labeled demonstrations to simulate human teachers, but it is part of the experimental setup, not part of our proposed protocol and algorithm.
Our agent does not have access to labeled demonstrations; it is initialized with \textit{random parameters} and is trained with only language-description feedback.
Last but not the least, we provide theoretical guarantees for $\mainalg$, while these works only present empirical studies.

\paragraph{Connection to Pragmatic Reasoning.} Another related line of research is work on the rational speech act (RSA) or pragmatic reasoning \citep{grice1975logic,golland-etal-2010-game,monroe2015pragmatics,goodman2016pragmatic,andreas-klein-2016-reasoning,fried-etal-2018-unified}, which is also concerned with transferring information via language. 
It is important to point out that RSA is a mental \textit{reasoning} model whereas \mainsetup is an \textit{interactive} protocol. 
In RSA, a speaker (or a listener) constructs a pragmatic message-encoding (or decoding) scheme by building an internal model of a listener (or a speaker). 
Importantly, during that process, one agent \textit{never} interacts with the other. 
In contrast, the \mainsetup agent learns through interaction with a teacher.
In addition, RSA focuses on encoding (or decoding) a single message while \mainsetup defines a process consisting of multiple rounds of message exchanging.  
We employ pragmatic inference to improve the quality of the simulated teachers but in our context, the technique is used to set up the experiments and is not concerned about communication between the teacher and the agent. 

\paragraph{Connection to Emergent Language.} Finally, our work also fundamentally differs from work on 
(RL-based) emergent language \citep{Foerster2016LearningTC,lazaridou2016multi,havrylov2017emergence,das2017learning,evtimova2017emergent,kottur-etal-2017-natural} in that we assume the teacher speaks a fixed, well-formed language, whereas in these works the teacher begins with  no language capability and learns a language over the course of training.

\section{Conclusion}

The communication protocol of a learning framework places natural boundaries on the learning efficiency of any algorithm that instantiates the framework.
In this work, we illustrate the benefits of designing learning algorithms based on a natural, descriptive communication medium like human language. 
Employing such expressive protocols leads to ample room for improving learning algorithms. 
Exploiting compositionality of language to improve sample efficiency, and learning with diverse types of feedback are interesting areas of future work.
Extending the theoretical analyses of \mainalg to more general settings is also an exciting open problem.

\section*{Acknowledgement}

We would like to thank Hal Daum\'e III, Kiant\'e Brantley, Anna Sotnikova, Yang Cao, Surbhi Goel, Akshay Krishnamurthy, and Cyril Zhang for their insightful comments on the paper. 
We thank Jacob Andreas for the discussion on pragmatic inference and thank Huyen Nguyen for useful conversations about human behavior. We also thank Microsoft GCR team for providing computational resources.

\bibliography{journal_full,paper,hanna,anthology,request_fulfilling}
\bibliographystyle{icml2021}


\appendix
\onecolumn

\section*{Appendix: Interactive Learning from Activity Description}

The appendix is organized as follows:
\begin{itemize}
    \item Statement and proof of theoretical guarantees for $\mainalg$ (\autoref{app:theory-analysis});
    \item Settings of the two problems we conduct experiments on (\autoref{app:app_problem});
    \item A practical implementation of the \mainalg algorithm that we use for experimentation (\autoref{sec:adel_practical});
    \item Training details including model architecture and hyperparameters (\autoref{app:training}); 
    \item Qualitative examples (\autoref{app:qual_example}).
\end{itemize}

We provide a list of notations in~\pref{tab:notations}.
\begin{table*}
\centering
\begin{tabular}{c|l}
    \midrule
    \textbf{Notation} & \textbf{Definition} \\
    \midrule
        $\Delta(\Ucal)$ & Space of all distributions over a set $\Ucal$\\ 
        $\unf(\Ucal)$ & Denotes the uniform distribution over a set $\Ucal$\\
        $\|.\|_p$ & $p$-norm\\
        $\KL{Q_1(x \mid y)}{Q_2(x \mid y)}$ & KL-divergence between two distributions $Q_1(\cdot \mid y)$ and $Q_2(\cdot \mid y)$ over a countable\\
        & set $\Xcal$. Formally, $\KL{Q_1(x \mid y)}{Q_2(x \mid y)} = \sum_{x \in \Xcal} Q_1(x \mid y) \ln \frac{Q_1(x \mid y)}{Q_2(x \mid y)}$. \\
        $\supp~Q(x)$ & Support of a distribution $Q \in \Delta(\Xcal)$. Formally, $\supp Q(x) = \{x \in \Xcal \mid Q(x) > 0\}$.\\
        $\NN$ & Set of natural numbers\\
    $\Scal$ & State space \\
    $ s$ & A single state in $\Scal$\\
    $\Acal$ & Finite action space \\
    $a$ & a single action in $\Acal$\\
    $\Dcal$ & Set of all possible descriptions and requests\\
    $d$ & A single description or request\\
    $T: \Scal \times \Acal \rightarrow \Delta(\Scal)$ & Transition function with $T(s' \mid s, a)$ denoting the probability of transitioning to \\
    & state $s'$ given state $s$ and action $a$. \\
    $\Rcal$ & Family of reward functions\\
    $R: \Scal \times \Acal \rightarrow [0, 1]$ & Reward function with $R(s, a)$ denoting the reward for taking action $a$ in state $s$ \\
    $H$ & Horizon of the problem denoting the number of actions in a single episode.\\
    $e$ & An execution $e = (s_1, a_1, s_2, \cdots, s_H, a_H)$ describing states and actions in an episode.\\
    $q=(R, d, s_1)$ & A single task comprising of reward function $R$, request $d$ and start state $s_1$ \\
    $\PP^\star(q)$ & Task distribution defined by the world\\
    $\PP^\star(e, R, s, d)$ & Joint distribution over executions and task (see~\pref{eqn:joint-dist}).\\
    $\PP_T(d \mid e)$ & Teacher model denoting distribution over descriptions $d$ for a given execution $e$. \\
    $\Theta$ & Set of all parameters of agent's policy.\\
    $\theta$ & Parameters of agent's policy. Belongs to the set $\Theta$.\\
    $\pi_\theta(a \mid s, d)$ & Agent's policy denoting the probability of action $a$ given state $s$, description $d$, \\
    & and parameters $\theta$.\\
    \bottomrule
\end{tabular}
\caption{List of common notations and their definitions.}
\label{tab:notations}
\end{table*}

\section{Theoretical Analysis of $\mainalg$}
\label{app:theory-analysis}

In this section, we provide a theoretical justification for an epoch-version of $\mainalg$ for the case of $H=1$. We prove consistency results showing $\mainalg$ learns a near-optimal policy, and we also derive the convergence rate under the assumption that we perform maximum likelihood estimation optimally and the teacher is consistent. We call a teacher model $\PP_T(d \mid e)$ to be consistent if for every execution $e$ and description $d$ we have $\PP_T(d \mid e) = \PP^\star(d \mid e)$. Recall that the conditional distribution $\PP^\star(d \mid e)$ is derived from the joint distribution defined  in~\pref{eqn:joint-dist}. We will use superscript $^\star$ to denote all probability distributions that are derived from this joint distribution.

We start by writing the epoch-version of $\mainalg$ in~\pref{alg:mainalg_epoch} for an arbitrary value of $H$. The epoch version of $\mainalg$ runs an outer loop of epochs (\pref{line:mainalg-for-loop-starts}-\ref{line:mainalg-epoch-for-loop-ends}). The agent model is updated only at the end of an epoch. In the inner loop (\pref{line:mainalg-epoch-inner-for-loop-starts}-\ref{line:mainalg-epoch-inner-for-loop-ends}), the agent samples a batch using the teacher model and the agent model. This is used to update the model at the end of the epoch. 

At the start of the $n^{th}$ epoch, our sampling scheme in~\pref{line:mainalg-epoch-sample-task}-\ref{line:mainalg-epoch-add-datapoint} defines a procedure to sample $(\hat{e}, \hat{d})$ from a distribution $D_n$ that remains fixed over this whole epoch. To define $D_n$, we first define $\PP_n(e) = \EE_{(R, d, s_1) \sim \PP^\star(q)}\left[\PP_{n}(e \mid s_1, d)\right]$ where we use the shorthand $\PP_n(e \mid s_1, d)$ to refer to $\PP_{\pi_{\theta_n}}(e \mid s_1, d)$. Note that $\hat{e} \sim \PP_n(e)$ in~\pref{line:mainalg-epoch-sample-traj}. As $\hat{d} \sim \PP^\star(d \mid \hat{e})$, therefore, we arrive at the following form of $D_n$:
\begin{equation}
 D_n(\hat{e}, \hat{d}) = \PP^\star(\hat{d} \mid \hat{e}) \PP_n(\hat{e}) \label{eqn:joint-distribution-epoch}. 
\end{equation}

We will derive our theoretical guarantees for $H=1$. This setting is known as the contextual bandit setting~\cite{Langford:07}, and while simpler than general reinforcement learning setting, it captures a large non-trivial class of problems. In this case, an execution $e = [s_1, a_1]$ can be described by the start state $s_1$ and a single action $a_1 \in \Acal$ taken by the agent. Since there is a single state and action in any execution, therefore, for cleaner notations we will drop the subscript and simply write $s, a$ instead of $s_1, a_1$. For convenience, we also define a few extra notations. Firstly, we define the marginal distribution $D_n(s, \hat{d}) = \sum_{a' \in \Acal} D_n([s, a'], d)$. Secondly, let $\PP^\star(s)$ be the marginal distribution over start state $s$ given by $\EE_{(R, d, s_1)  \sim \PP^\star(q)}[\one\{s_1=s\}]$. We state some useful relations between these probability distributions in the next lemma.

\begin{algorithm}[H]
\centering
\caption{\small $\epochmainalg$: Epoch Version of $\mainalg$. We assume the teacher is consistent, i.e., $\PP_T(d \mid e) = \PP^\star(d \mid e)$ for every $(d, e)$.}
\label{alg:mainalg_epoch}
\begin{algorithmic}[1]
\State \textbf{Input}: teacher model $\PP^\star(d \mid e)$ and task distribution model $\PP^\star(q)$.
\State Initialize agent policy $\pi_{\theta_1}: \Scal \times \Dcal \rightarrow \unf(\Acal)$ \label{line:mainalg-epoch-policy-init}
\For{$n = 1, 2, \cdots, N$} \label{line:mainalg-epoch-for-loop-starts}
\State $\Bcal = \emptyset$ \label{line:mainalg-epoch-dataset-init}
\For{$m=1, 2, \cdots, M$} \label{line:mainalg-epoch-inner-for-loop-starts}
\State World samples $q=(R, d^{\star}, s_1) \sim \PP^{\star}(\cdot)$ \label{line:mainalg-epoch-sample-task}
\State Agent generates $\hat{e} \sim \PP_{\pi_{\theta_n}}(\cdot \mid s_1, d^{\star})$ \label{line:mainalg-epoch-sample-traj}
\State Teacher generates description $\hat{d} \sim \PP^\star(\cdot \mid \hat{e})$  \label{line:mainalg-epoch-sample-description}
\State $\Bcal \leftarrow \Bcal \cup \left\{\left(\hat{e}, \hat{d}\right)\right\}$ \label{line:mainalg-epoch-add-datapoint}
\EndFor \label{line:mainalg-epoch-inner-for-loop-ends}
\State Update agent policy using batch updates:
\begin{align}
    \theta_{n+1} \leftarrow \arg\max_{\theta' \in \Theta} \sum_{(\hat{e}, \hat{d}) \in \Bcal} \sum_{(s, a_s) \in \hat{e}} \log \pi_{\theta'}(a_s \mid s, \hat{d}) \nonumber
\end{align} \par where $a_s$ is the action taken by the agent in state $s$ in execution $\hat{e}$. \label{line:mainalg-epoch-update}
\EndFor \label{line:mainalg-epoch-for-loop-ends}
\Return $\pitheta$
\end{algorithmic}
\end{algorithm}

\begin{lemma}\label{lem:useful-lemmas} For any $n \in \NN$, we have:
\begin{equation}\label{eqn:useful-prior-property}
    \PP_n(e \defeq [s, a]) = \PP^\star(s) \PP_n(a \mid s), \mbox{ where } \PP_n(a \mid s) \defeq \sum_{d} \PP^\star(d \mid s) \PP_n(a \mid s, d).
\end{equation}
\end{lemma}
\begin{proof} We first compute the marginal distribution $\sum_{a' \in \Acal} \PP_n(e' \defeq [s, a'])$ over $s$:
\begin{equation*}
    \sum_{a' \in \Acal} \PP_n(e' \defeq [s, a']) = \sum_{a' \in \Acal}\sum_{R, d} \PP^\star(R, d, s) \PP_n(a' \mid s, d) = \sum_{R, d} \PP^\star(R, d, s) = \PP^\star(s).
\end{equation*}
Next we compute the conditional distribution $\PP_n(a \mid s)$ as shown:
\begin{align*}
    \PP_n(a \mid s) = \frac{\PP_n([s, a])}{\sum_{a' \in \Acal} \PP_n([s, a'])} &= \sum_{R, d}\frac{\PP^\star(R, d, s) \PP_n(a \mid s, d) }{\PP^\star(s)} =  \sum_{d} \frac{\PP^\star(s, d) \PP_n(a \mid s, d) }{\PP^\star(s)}= \sum_{d} \PP^\star(d \mid s) \PP_n(a \mid s, d).
\end{align*}
This also proves $\PP_n([s, a]) = \PP^\star(s)\PP_n(a \mid s)$.
\end{proof}

For $H=1$, the update equation in~\pref{line:mainalg-epoch-update} solves the following optimization equation:
\begin{equation}
    \max_{\theta' \in \Theta} J_n(\theta) \qquad \mbox{where }~~J_n(\theta) \defeq \sum_{ (\hat{e} \defeq [s, a], \hat{d}) \in \Bcal} \ln \pi_{\theta'}(a \mid s, \hat{d}).
\end{equation} 

Here $J_n(\theta)$ is the empirical objective whose expectation over draws of batches is given by:
\begin{equation*}
    \EE[J_n(\theta)] = \EE_{(\hat{e}=[s, a], d) \sim D_n} \left[ \ln \pi_{\theta}(a \mid s, d)\right].
\end{equation*}

As this is negative of the cross entropy loss, the Bayes optimal value would be achieved for $\pi_{\theta}(a \mid s, d) = D_n(a \mid s, d)$ for all $a \in \Acal$ and every $(s, d) \in \supp D_n(s, d)$. We next state the form of this Bayes optimal model and then state our key realizability assumption.

\begin{lemma}\label{lem:bayes-optimal-form} Fix $n \in \NN$. For every $(s, d) \in \supp D_n(s, d)$ the value of the Bayes optimal model $D_n(a \mid s, d)$ at the end of the $n^{th}$ epoch is given by:
\begin{equation*}
    D_n(a \mid s, d) = \frac{\PP^\star(d \mid [s, a]) \PP_n(a \mid s)}{\sum_{a' \in \Acal} \PP^\star(d \mid [s, a']) \PP_n(a' \mid s)}. 
\end{equation*}
\end{lemma}
\begin{proof}
The Bayes optimal model is given by $D_n(a \mid s, d)$ for every $(s, d) \in \supp D_n(s, d)$. We compute this using Bayes' theorem.
\begin{align*}
   D_n(a \mid s, d) = \frac{D_n([s, a], d)}{\sum_{a' \in \Acal}D_n([s, a'], d)} = \frac{\PP^\star(d \mid [s, a]) \PP_n([s, a])}{\sum_{a' \in \Acal}\PP^\star(d \mid [s, a']) \PP_n([s, a'])} = \frac{\PP^\star(d \mid [s, a]) \PP_n(a \mid s)}{\sum_{a' \in \Acal}\PP^\star(d \mid [s, a']) \PP_n(a' \mid s)}.
\end{align*}
The last equality above uses~\pref{lem:useful-lemmas}.
\end{proof}

In order to learn the Bayes optimal model, we need our policy class to be expressive enough to contain this model. We formally state this \emph{realizability} assumption below.
\begin{assumption}[Realizability]\label{assum:realizability} For every $\theta \in \Theta$, there exists $\theta' \in \Theta$ such that for every start state $s$, description $d$ we have:
\begin{equation*}
\forall a \in \Acal, \quad   \pi_{\theta'}(a \mid s, d) = \frac{\PP^\star(d \mid [s, a]) Q_\theta(a \mid s)}{\sum_{a' \in \Acal} \PP^\star(d \mid [s, a']) Q_\theta(a' \mid s)}, \quad \mbox{where} \quad Q_\theta(a \mid s) = \sum_{d'} \PP^\star(d' \mid s) \pi_\theta(a \mid s, d'). 
\end{equation*}
\end{assumption}

We can use the realizability assumption along with convergence guarantees for log-loss to state the following result:
\begin{theorem}[Theorem 21 of~\cite{agarwal2020flambe}]\label{lem:generalization-bound} Fix $m \in \NN$ and $\delta \in (0, 1)$. Let $\{(d^{(i)}, e^{(i)}=[s^{(i)}, a^{(i)}]\}_{i=1}^m$ be i.i.d draws from $D_n(e, d)$ and let $\theta_{n+1}$ be the solution to the optimization problem in~\pref{line:mainalg-epoch-update} of the $n^{th}$ epoch of $\epochmainalg$. Then with probability at least $1-\delta$ we have:
\begin{equation}
    \EE_{s, d \sim D_n}\left[\|D_{n}(a \mid s, d) - \PP_{\pi_{\theta_{n+1}}}(a \mid s, d)\|_1 \right] \le C\sqrt{\frac{1}{m}\ln \nicefrac{|\Theta|}{\delta}},
\end{equation}
where $C > 0$ is a universal constant.
\end{theorem}
Please see~\citet{agarwal2020flambe} for a proof. \pref{lem:generalization-bound} implies that assuming realizability, as $M \rightarrow \infty$, our learned solution converges to the Bayes optimal model pointwise on the support over $D_n(s, d)$. Since we are only interested in consistency, we will assume $M \rightarrow \infty$ and assume $\PP_{n+1}(a \mid s, d) = D_n(a \mid s, d)$ for every $(s, d) \in \supp D_n(s, d)$. We will refer to this as optimally performing the maximum likelihood estimation at $n^{th}$ epoch. If the learned policy is given by $\PP_{n+1}(a \mid s, d) = D_n(a \mid s, d)$, then the next Lemma states the relationship between the marginal distribution $\PP_{n+1}(a \mid s)$ for the next time epoch and marginal $\PP_n(a \mid s)$ for this epoch.

\begin{lemma}[Inductive Relation Between Marginals]\label{lem:inductive-relation-marginals} For any $n \in \NN$, if we optimally perform the maximum likelihood estimation at the $n^{th}$ epoch of $\epochmainalg$, then for all start states $s$, the marginal distribution $\PP_{n+1}(a \mid s)$ for the $(n+1)^{th}$ epoch is given by:
\begin{equation*}
    \PP_{n+1}(a \mid s) = \sum_{d}  \frac{\PP^\star(d \mid [s, a]) \PP_n(a \mid s)\PP^\star(d \mid s)}{\sum_{a' \in \Acal}\PP^\star(d \mid [s, a']) \PP_n(a' \mid s)}.
\end{equation*}
\end{lemma}
\begin{proof} The proof is completed as follows:
\begin{equation*}
    \PP_{n+1}(a \mid s) = \sum_{d} \PP^\star(d \mid s) \PP_{n+1}(a \mid s, d) = \sum_{d}  \frac{\PP^\star(d \mid [s, a]) \PP_n(a \mid s)\PP^\star(d \mid s)}{\sum_{a' \in \Acal}\PP^\star(d \mid [s, a']) \PP_n(a' \mid s)},
\end{equation*}
where the first step uses~\pref{lem:useful-lemmas} and the second step uses $\PP_{n+1}(a \mid s, d) = D_n(a \mid s, d)$ (optimally solving maximum likelihood) and the form of $D_n$ from~\pref{lem:bayes-optimal-form}.
\end{proof}

\subsection{Proof of Convergence for Marginal Distribution}

Our previous analysis associates a probability distribution $\PP_n(a \mid s, d)$ and $\PP_n(a \mid s)$ with the $n^{th}$ epoch of $\epochmainalg$. For any $n \in \NN$, the $n^{th}$ epoch of $\epochmainalg$ can be viewed as a transformation of $\PP_n(a \mid s, d) \mapsto \PP_{n+1}(a \mid s, d)$ and $\PP_n(a \mid s) \mapsto \PP_{n+1}(a \mid s)$. In this section, we show that under certain conditions, the running average of the marginal distributions $\PP_n(a \mid d)$ converges to the optimal marginal distribution $\PP^\star(a \mid d)$. We then discuss how this can be used to learn the optimal policy $\PP^\star(a \mid s, d)$.

We use a potential function approach to measure the progress of each epoch. Specifically, we will use KL-divergence as our choice of potential function. The next lemma bounds the change in potential after a single iteration.

\begin{lemma}\label{lem:potential-difference}[Potential Difference Lemma] For any $n \in \NN$ and start state $s$, we define the following distribution over descriptions $\PP_n(d \mid s) \defeq \sum_{a' \in \Acal} \PP^\star(d \mid [s, a]) \PP_n(a \mid s)$. Then for every start state $s$ we have:
\begin{equation*}
    \KL{\PP^\star(a \mid s)}{\PP_{n+1}(a \mid s)} - \KL{\PP^\star(a \mid s)}{\PP_{n}(a \mid s)} \le - \KL{\PP^\star(d \mid s)}{\PP_n(d \mid s)}.
\end{equation*}
\end{lemma}
\begin{proof} The change in potential from the start of $n^{th}$ epoch to its end is given by:
\begin{equation}\label{eqn:potential-difference}
    \KL{\PP^\star(a \mid s)}{\PP_{n+1}(a \mid s)} - \KL{\PP^\star(a \mid s)}{\PP_{n}(a \mid s)} = - \sum_{a \in \Acal}  \PP^\star(a \mid s) \ln \left( \frac{\PP_{n+1}(a \mid s)}{\PP_{n}(a \mid s)} \right)
\end{equation}

Using~\pref{lem:inductive-relation-marginals} and the definition of $\PP_n(d \mid s)$ we get:
\begin{equation*}
    \frac{\PP_{n+1}(a \mid s)}{\PP_n(a \mid s)} = \sum_{d} \frac{\PP^\star(d \mid [s, a])\PP^\star(d \mid s)}{\sum_{a' \in \Acal} \PP^\star(d \mid [s, a']) \PP_n(a' \mid s)} = \sum_{d} \PP^\star(d \mid [s, a])\frac{\PP^\star(d \mid s)}{\PP_n(d \mid s)}.
\end{equation*}
Taking logarithms and applying Jensen's inequality gives:
\begin{equation}
    \ln \left(\frac{\PP_{n+1}(a \mid s)}{\PP_{n}(a \mid s)}\right) = \ln\left( \sum_{d} \PP^\star(d \mid [s, a])\frac{\PP^\star(d \mid s)}{\PP_n(d \mid s)}\right) \ge \sum_{d} \PP^\star(d \mid [s, a]) \ln \left(\frac{\PP^\star(d \mid s)}{\PP_n(d \mid s)}\right).
\end{equation}

Taking expectations of both sides with respect to $\PP^\star(a \mid s)$ gives us:
\begin{align*}
    \sum_{a} \PP^\star(a \mid s) \ln \left(\frac{\PP_{n+1}(a \mid s)}{\PP_{n}(a \mid s)}\right) &\ge \sum_{a} \sum_{d}  \PP^\star(a \mid s)  \PP^\star(d \mid [s, a]) \ln \left(\frac{\PP^\star(d \mid s)}{\PP_n(d \mid s)}\right) \\
    &= \sum_{d} \PP^\star(d \mid s) \ln \left(\frac{\PP^\star(d \mid s)}{\PP_n(d \mid s)}\right)\\
    &= \KL{\PP^\star(d \mid s)}{\PP_n(d \mid s)}
\end{align*}

where the last step uses the definition of $\PP_n(d \mid s)$. The proof is completed by combining the above result with~\pref{eqn:potential-difference}.
\end{proof}

\paragraph{The $\PP_{s}$ matrix.} For a fixed start state $s$, we define $\PP_s$ as the matrix whose entries are $\PP^\star(d \mid [s, a])$. The columns of this matrix range over actions, and the rows range over descriptions. We denote the minimum singular value of the description matrix $\PP_s$ by $\smin(s)$. 

We state our next assumption that the minimum singular value of $\PP_s$ matrix is non-zero.
\begin{assumption}[Minimum Singular Value is Non-Zero]\label{assum:singular-values} For every start state $s$, we assume $\smin(s) > 0$.
\end{assumption}
Intuitively, this assumption states that there is enough information in the descriptions for the agent to decipher probabilities over actions from learning probabilities over descriptions. More formally, we are trying to decipher $\PP^\star(a \mid s)$ using access to two distributions: $\PP^\star(d \mid s)$ which generates the initial requests, and the teacher model $\PP^\star(d \mid [s, a])$ which is used to describe an execution $e = [s, a]$. This can result in an underspecified problem. The only constraints these two distributions place on $\PP^\star(a \mid s)$ is that $\sum_{a \in \Acal} \PP^\star(d \mid [s, a]) \PP^\star(a \mid s) = \PP^\star(d \mid s)$. This means all we know is that $\PP^\star(a \mid s)$ belongs to the following set of solutions of the previous linear systems of equation: \begin{equation*}
\left\{Q(a \mid s) \mid \sum_{a \in \Acal} \PP^\star(d \mid [s, a]) Q(a \mid s) = \PP^\star(d \mid s)~\forall d, \qquad Q(a \mid s) \mbox{ is a distribution}\right\}.
\end{equation*}
As $\PP^\star(a \mid s)$ belongs to this set hence this set is nonempty. However, if we also assume that $\smin(s) > 0$ then the above set has a unique solution. Recall that singular values are square root of eigenvalues of $\PP^\top_{s} \PP_s$, and so $\smin(s) > 0$ implies that the matrix $\PP^\top_{s} \PP_s$ is invertible. \footnote{Recall that a matrix of the form $A^\top A$ always have non-negative eigenvalues.} This means, we can find the unique solution of the linear systems of equation by multiplying both sides by $(\PP^\top_{s}\PP_{s})^{-1}\PP^\top_{s}$. Hence,~\pref{assum:singular-values} makes it possible for us to find $\PP^\star(a \mid s)$ using just the information we have. Note that we cannot solve the linear system of equations directly since the description space and action space can be extremely large. Hence, we use an oracle based solution via reduction to supervised learning.

The next theorem shows that the running average of learned probabilities $\PP_n(a \mid s)$ converges to the optimal marginal distribution $\PP^\star(a \mid s)$ at a rate determined by the inverse square root of the number of epochs of $\mainalg$, the minimum singular value of the matrix $\PP_{s}$, and the KL-divergence between optimal marginal and initial value.
\begin{theorem}\label{thm:convergence-rate-marginal}[Rate of Convergence for Marginal] For any $t \in \NN$ we have:
\begin{equation*}
     \|\PP^\star(a \mid s) - \frac{1}{t} \sum_{n=1}^{t} \PP_n(a \mid s)\|_2 \le \frac{1}{\smin(s)}\sqrt{\frac{2}{ t} \KL{\PP^\star(a \mid s)}{\PP_1(a \mid s)}},
\end{equation*}
and if $\PP_1(a \mid s, d)$ is a uniform distribution for every $s$ and $d$, then \begin{equation*}
     \|\PP^\star(a \mid s) - \frac{1}{t} \sum_{n=1}^{t} \PP_n(a \mid s)\|_2 \le \frac{1}{\smin(s)}\sqrt{\frac{2\ln|\Acal|}{t}}.
\end{equation*}
\end{theorem}
\begin{proof}
We start with~\pref{lem:potential-difference} and bound the right hand side as shown:
\begin{align*}
    \KL{\PP^\star(a \mid s)}{\PP_{n+1}(a \mid s)} - \KL{\PP^\star(a \mid s)}{\PP_{n}(a \mid s)} &\le - \KL{\PP^\star(d \mid s)}{\PP_n(d \mid s)}\\
    &\le - \frac{1}{2}\|\PP^\star(d \mid s) - \PP_n(d \mid s)\|^2_1, \\
    &\le - \frac{1}{2}\|\PP^\star(d \mid s) - \PP_n(d \mid s)^2\|^2_2 \\
    &= - \frac{1}{2} \| \PP_{s} \left\{\PP^\star(a \mid s) - \PP_n(a \mid s)\right\} \|^2_2, \\
    &\le - \frac{1}{2} \smin(s)^2 \|\PP^\star(a \mid s) - \PP_n(a \mid s)\|^2_2,
\end{align*}
where the second step uses Pinsker's inequality. The third step uses the property of $p$-norms, specifically, $\|\nu\|_2 \le \|\nu\|_1$ for all $\nu$. The fourth step, uses the definition of $\PP^\star(d \mid s) = \sum_{a' \in \Acal} \PP(d \mid s, a')\PP^\star(a' \mid s))$ and $\PP_n(d \mid s) = \sum_{a' \in \Acal} \PP(d \mid s, a')\PP_n(a' \mid s)$. We interpret the notation $\PP^\star(a \mid s)$ as a vector over actions whose value is the probability $\PP^\star(a \mid s)$. Therefore, $\PP_s \PP^\star(a \mid s)$ represents a matrix-vector multiplication. Finally, the last step, uses $\|Ax\|_2 \ge \smin(A) \|x\|_2$ for any vector $x$ and matrix $A$ of compatible shape such that $Ax$ is defined, where $\smin(A)$ is the smallest singular value of $A$.

Summing over $n$ from $n=1$ to $t$ and rearranging the terms we get:
\begin{equation*}
    \KL{\PP^\star(a \mid s)}{\PP_{t+1}(a \mid s)} \le \KL{\PP^\star(a \mid s)}{\PP_1(a \mid s)} - \frac{1}{2} \smin(s)^2 \sum_{n=1}^{t} \|\PP^\star(a \mid s) - \PP_n(a \mid s)\|^2_2.
\end{equation*}
As the left hand-side is positive we get:
\begin{equation*}
    \sum_{n=1}^{t} \|\PP^\star(a \mid s) - \PP_n(a \mid s)\|^2_2 \le \frac{2}{\smin(s)^2} \KL{\PP^\star(a \mid s)}{\PP_1(a \mid s)}.
\end{equation*}
Dividing by $t$ and applying Jensen's inequality (specifically, $\EE[X^2] \ge \EE[|X|]^2$) we get:
\begin{equation}
    \frac{1}{t} \sum_{n=1}^{t} \|\PP^\star(e) - \PP_n(e)\|_2 \le \frac{1}{\smin(s)}\sqrt{\frac{2}{ t} \KL{\PP^\star(a \mid s)}{\PP_1(a \mid s)}}
\end{equation}
Using the triangle inequality, the left hand side can be bounded as:
\begin{equation}
    \frac{1}{t} \sum_{n=1}^{t} \|\PP^\star(a \mid s) - \PP_n(a \mid s)\|_2 \ge \|\PP^\star(a \mid s) - \frac{1}{t} \sum_{n=1}^{t} \PP_n(a \mid s)\|_2
\end{equation}
Combining the previous two equations proves the main result. Finally, note that if $\PP_1(a \mid s, d) = \nicefrac{1}{|\Acal|}$ for every value of $s, d, $ and $a$, then $\PP_1(a \mid s)$ is also a uniform distribution over actions. The initial KL-divergence is then bounded by $\ln |\Acal|$ as shown below:
\begin{equation*}
    \KL{\PP^\star(a \mid s)}{\PP_1(a \mid s)} = - \sum_{a \in \Acal} \PP^\star(a \mid s) \ln \frac{1}{|\Acal|} + \sum_{a \in \Acal} \PP^\star(a \mid s) \ln \PP^\star(a \mid s) \le \ln |\Acal|,
\end{equation*}
where the second step uses the fact that entropy of a distribution is non-negative. This completes the proof.
\end{proof}

\subsection{Proof of Convergence to Near-Optimal Policy}

Finally, we discuss how to learn $\PP^\star(a \mid s, d)$ once we learn $\PP^\star(a \mid s)$. Since we only derive convergence of running average of $\PP_n(a \mid s)$ to $\PP^\star(a \mid s)$, therefore, we cannot expect $\PP_n(a \mid s, d)$ to converge to $\PP^\star(a \mid s, d)$. Instead, we will show that if we perform~\pref{line:mainalg-epoch-dataset-init}-\ref{line:mainalg-epoch-update} in~\pref{alg:mainalg_epoch} using the running average of policies, then the learned Bayes optimal policy will converge to the near-optimal policy. The simplest way to accomplish this with~\pref{alg:mainalg_epoch} is to perform the block of code in~\pref{line:mainalg-epoch-dataset-init}-\ref{line:mainalg-epoch-update} twice, once when taking actions according to $\PP_n(a \mid s, d)$, and once when taking actions according to running average policy $\tilde{\PP}_n(a \mid s, d) = \frac{1}{n}\sum_{t=1}^n \tilde{\PP}_t(a \mid s, d)$. This will give us two Bayes optimal policy in~\ref{line:mainalg-epoch-update} one each for the current policy $\PP_n(a \mid s, d)$ and the running average policy $\tilde{\PP_n}(a \mid s, d)$. We use the former for roll-in in the future and the latter for evaluation on held-out test set.

For convenience, we first define an operator that denotes mapping of one agent policy to another.

\paragraph{$W$ operator.} Let $\PP(a \mid s, d)$ be an agent policy used to generate data in any epoch of $\epochmainalg$ (\pref{line:mainalg-epoch-inner-for-loop-starts}-\ref{line:mainalg-epoch-add-datapoint}). We define the $W$ operator as the mapping to the Bayes optimal policy for the optimization problem solved by $\epochmainalg$ in~\pref{line:mainalg-epoch-update} which we denote by $(W\PP)$. Under the realizability assumption (\pref{assum:realizability}), the agent learns the $W \PP$ policy when $M \rightarrow \infty$. Using~\pref{lem:useful-lemmas} and ~\pref{lem:bayes-optimal-form}, we can verify that:
\begin{equation*}
    (W\PP)(a \mid s, d) = \frac{\PP^\star(d \mid [s, a]) \PP(a \mid s)}{\sum_{a' \in \Acal} \PP^\star(d \mid [s, a']) \PP(a' \mid s)}, \quad \mbox{where} \quad \PP(a \mid s) = \sum_{d} \PP^\star(d \mid s) \PP(a \mid s, d).
\end{equation*}

We first show that our operator is smooth around $\PP^\star(a \mid s)$. 
\begin{lemma}[Smoothness of $W$]\label{lem:smoothness-operator} For any start state $s$ and description $d \in \supp~\PP^\star(d \mid s)$, there exists a finite constant $K_s$ such that:
\begin{equation*}
    \| W\PP(a \mid s, d) - W\PP^\star(a \mid s, d)\|_1 \le K_s \|\PP(a \mid s) - \PP^\star(a \mid s)\|_1.
\end{equation*}
\end{lemma}
\begin{proof} We define $\PP(d \mid s) = \sum_{a' \in \Acal} \PP^\star(d \mid s, a') \PP(a' \mid s)$. Then from the definition of operator $W$ we have:
\begin{align*}
    & \left| W\PP(a \mid s, d) - W\PP^\star(a \mid s, d) \right|_1 \\
    &= \sum_{a \in \Acal} \left| \frac{\PP^\star(d \mid [s, a])\PP(a \mid s)}{\PP(d \mid s)}  - \frac{\PP^\star(d \mid [s, a])\PP^\star(a \mid s)}{\PP^\star(d \mid s)}  \right|\\
    &= \sum_{a \in \Acal} \PP^\star(d \mid [s, a])\frac{ \left| \PP(a \mid s)\PP^\star(d \mid s) - \PP^\star(a \mid s) \PP(d \mid s)\right|}{\PP(d \mid s)\PP^\star(d \mid s)}\\
    &\le \sum_{a \in \Acal} \PP^\star(d \mid [s, a])\PP(a \mid s) \frac{ \left| \PP^\star(d \mid s) - \PP(d \mid s)\right|}{\PP(d \mid s)\PP^\star(d \mid s)} + \sum_{a \in \Acal} \PP^\star(d \mid [s, a])\frac{ \left| \PP(a \mid s) - \PP^\star(a \mid s) \right|}{\PP^\star(d \mid s)}\\
    &= \frac{ \left| \PP^\star(d \mid s) - \PP(d \mid s)\right|}{\PP^\star(d \mid s)} + \sum_{a \in \Acal} \PP^\star(d \mid [s, a])\frac{ \left| \PP(a \mid s) - \PP^\star(a \mid s) \right|}{\PP^\star(d \mid s)}\\
    &\le 2\sum_{a \in \Acal} \PP^\star(d \mid [s, a])\frac{ \left| \PP(a \mid s) - \PP^\star(a \mid s) \right|}{\PP^\star(d \mid s)}, \qquad \mbox{(using the definition of $\PP(d\mid s)$)} \\
    &\le \frac{2}{\PP^\star(d \mid s)}\| \PP(a \mid s) - \PP^\star(a \mid s) \|_1.
\end{align*}
Note that the policy will only be called on a given pair of $(s, d)$ if and only if $\PP^\star(d \mid s) > 0$, hence, the constant is bounded. We define $K_s = \max_{d} \frac{2}{\PP^\star(d \mid s)}$ where maximum is taken over all descriptions $d \in \supp~\PP^\star(d \mid s)$.
\end{proof}

\begin{theorem}[Convergence to Near Optimal Policy] Fix $t \in \NN$, and let $\tilde{\PP}_t(a \mid s, d) = \frac{1}{t} \sum_{n=1}^t \PP_n(a \mid s, d)$ be the average of the agent's policy across epochs. Then for every start state $s$ and description $d \in \supp~\PP^\star(d \mid s)$ we have:
 \begin{equation*}
    \lim_{t \rightarrow \infty} (W \tilde{\PP}_t)(a \mid s, d) = \PP^\star(a \mid s, d).
\end{equation*}
\end{theorem}
\begin{proof} Let $\tilde{\PP}_t(a \mid s) = \sum_d \PP^\star(d \mid s) \tilde{\PP}_t(a \mid s, d)$. Then it is easy to see that  $\tilde{\PP}_t(a \mid s) = \frac{1}{t} \sum_{n=1}^t \PP_n(a \mid s)$. From~\pref{thm:convergence-rate-marginal} we have $\lim_{t \rightarrow \infty} \|\tilde{\PP}_t(a \mid s) - \PP^\star(a \mid s)\|_2 = 0$. As $\Acal$ is finite dimensional, therefore, $\|\cdot \|_2$ and $\|\cdot\|_1$ are equivalent, i.e., convergence in one also implies convergence in the other. This implies, $\lim_{t \rightarrow \infty} \|\tilde{\PP}_t(a \mid s) - \PP^\star(a \mid s)\|_1 = 0$. 

From~\pref{lem:smoothness-operator} we have:
\begin{equation*}
    \lim_{t \rightarrow \infty} \|(W \tilde{\PP}_t)(a \mid s, d) - (W \PP^\star)(a \mid s, d) \|_1 \le K_s \lim_{t \rightarrow \infty}  \| \tilde{\PP}_t(a \mid s) - \PP^\star(a \mid s)\|_1 = 0.
\end{equation*}

This shows $\lim_{t \rightarrow \infty}(W \tilde{\PP}_t)(a \mid s, d) =  (W \PP^\star)(a \mid s, d)$. Lastly, we show that the optimal policy $\PP^\star(a \mid s, d)$ is a fixed point of $W$:
\begin{equation*}
    (W \PP^\star)(a \mid s, d) =  \frac{\PP^\star(d \mid s, a) \PP^\star(a \mid s)}{\sum_{a' \in \Acal} \PP^\star(d \mid s, a') \PP^\star(a' \mid s)} = \frac{\PP^\star(d, a \mid s)}{\sum_{a' \in \Acal} \PP^\star(d, a' \mid s)} = \frac{\PP^\star(d, a \mid s)}{\PP^\star(d \mid s)} = \PP^\star(a \mid s, d).
\end{equation*}
This completes the proof.
\end{proof}

\section{Problem settings}
\label{app:app_problem}

\begin{figure}[t!]
    \centering
     \begin{subfigure}{0.52\textwidth}
         \centering
         \includegraphics[width=0.9\textwidth]{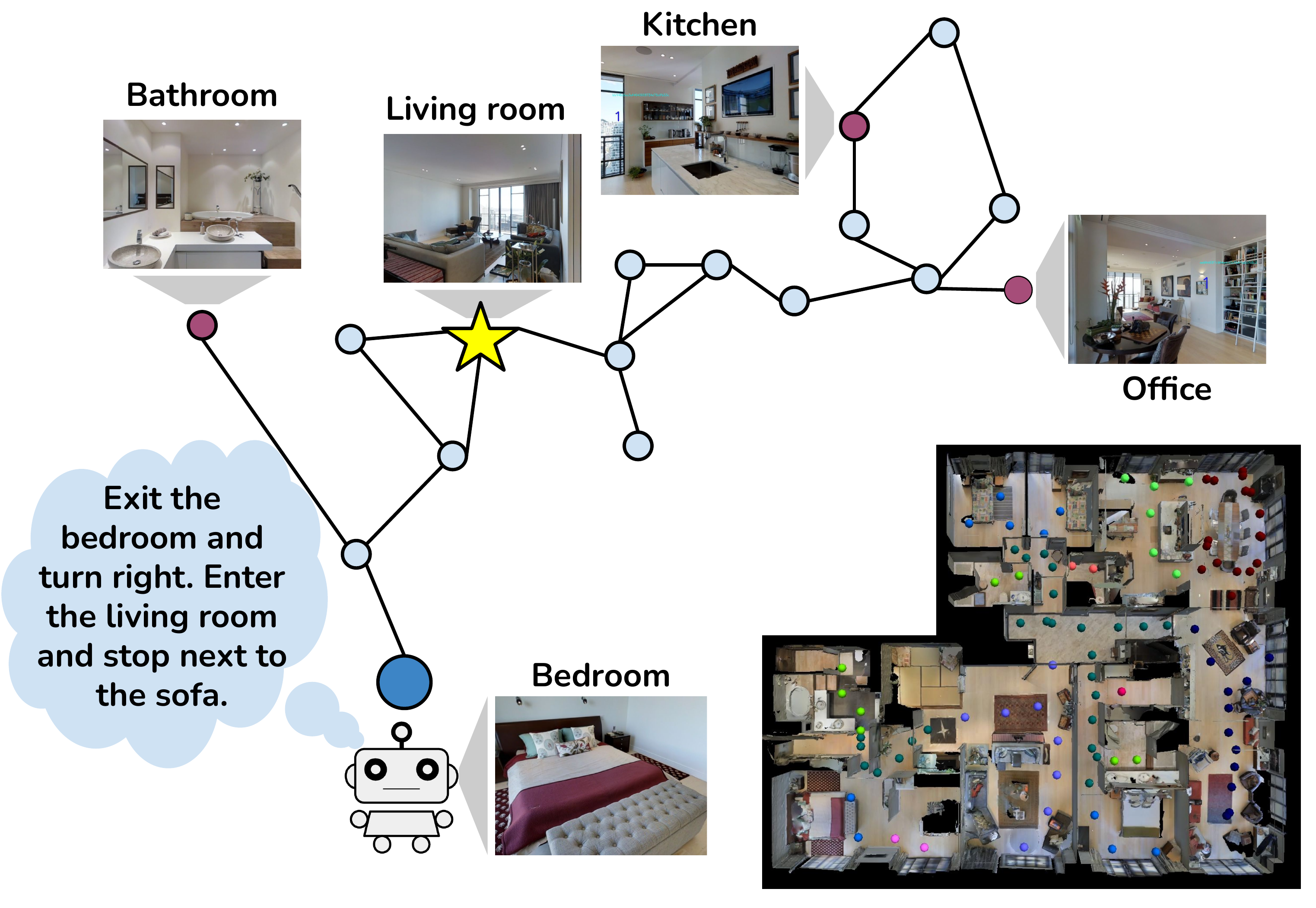}
         \caption{\textit{Vision-language navigation} (\nav): a (robot) agent fulfills a navigational natural-language request in a photo-realistic simulated house. Locations in the house are connected as a graph. In each time step, the agent receives a photo of the panoramic view at its current location (due to space limit, here we only show part of a view). Given the view and the language request, the agent chooses an adjacent location to go to. On average, each house has about 117 locations.  }
         \label{fig:nav_task}
     \end{subfigure}
     \hfill
     \begin{subfigure}{0.45\textwidth}
         \centering
         \includegraphics[width=\textwidth]{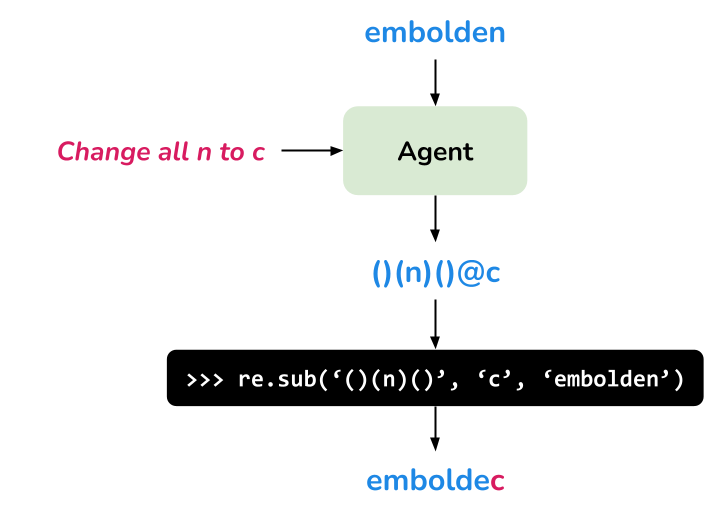}
         \caption{\textit{Word modification} (\regex): an agent is given an input word and a natural-language request that asks it to modify the word. The agent outputs a regular expression that follows our specific syntax. The regular expression is executed by the Python's \texttt{re.sub()} method to generate an output word. }
         \label{fig:regex_task}
     \end{subfigure}
     \caption{Illustrations of the two request-fulfilling problems that we conduct experiments on.}
     \label{fig:example}
\end{figure}

\autoref{fig:example} illustrates the two problems that we conduct experiments on.

\subsection{Vision-Language Navigation}
\label{app:app_vln_setup}

\paragraph{Environment Simulator and Data.} We use the Matterport3D simulator and the Room-to-Room dataset\footnote{\url{https://github.com/peteanderson80/Matterport3DSimulator/blob/master/tasks/R2R/data/download.sh}} developed by \citet{anderson2018vision}.
The simulator photo-realistically emulates the first-person view of a person walking in a house.
The dataset contains tuples of human-generated English navigation requests annotated with ground-truth paths in the environments.
To evaluate on the test set, the authors require submitting predictions to an evaluation site\footnote{\url{https://eval.ai/web/challenges/challenge-page/97/overview}}, which limits the number of submissions to five.
As our goal is not to establish state-of-the-art results on this task, but to compare performance of multiple learning frameworks, we re-split the data into 4,315 simulation, 2,100 validation, and 2,349 test data points. 
The simulation split, which is used to simulate the teacher, contains three requests per data point (i.e. $|\Dcal_n^{\star}| = 3$). 
The validation and test splits each contains only one request per data point. 
On average, each request includes 2.5 sentences and 26 words.
The word vocabulary size is 904 and the average number of optimal actions required to reach the goal is 6.

\paragraph{Simulated Teacher.}
We use SDTW \citep{magalhaes2019general} as the \texttt{perf} metric and set the threshold $\tau = 0.5$.
The SDTW metric re-weights success rate by the shortest (order-preserving) alignment distance between a predicted path and a ground-truth path, offering more fine-grained evaluation of navigation paths. 

\paragraph{Approximate marginal $P_{\piomega}(e \mid s_1)$.} 
The approximate marginal is a function that takes in a start location $s_1$ and randomly samples a shortest path on the environment graph that starts at $s_1$ and has (unweighted) length between 2 and 6.  

\subsection{Word Modification}
\label{app:app_regex_setup}

\paragraph{Regular Expression Compiler.} We use Python 3.7's \href{https://docs.python.org/3.9/library/re.html#re.sub}{\texttt{re.sub(pattern, replace, string)}} method as the regular expression compiler. 
The method replaces every substring of \texttt{string} that matches a regular expression \texttt{pattern} with the string \texttt{replace}.
A regular expression predicted by our agent $\hat{a}_{1:H}$ has the form ``\texttt{pattern@replace}", where \texttt{pattern} and \texttt{replace} are strings and \texttt{@} is the at-sign character. 
For example, given the word \textit{embolden} and the request ``\textit{replace all n with c}", the agent should ideally generate the regular expression ``\texttt{()(n)()@c}".
We then split the regular expression by the character \texttt{@} into a string $\texttt{pattern} = ``\texttt{()(n)()}"$ and a string $\texttt{replace} = ``\textit{c}"$.
We execute the Python's command \texttt{re.sub(`()(n)()', `c', `embolden')} to obtain the output word \textit{emboldec}.

\paragraph{Data.} We use the data collected by \citet{andreas-etal-2018-learning}.
The authors presented crowd-workers with pairs of input and output words where the output words are generated by applying regular expressions onto the input words.
Workers are asked to write English requests that describe the change from the input words to the output words. 
From the human-generated requests, the authors extracted 1,917 request templates.
For example, a template has the form \textit{add an \texttt{AFTER} to the start of words beginning with \texttt{BEFORE}}, where \texttt{AFTER} and \texttt{BEFORE} can be replaced with latin characters to form a request. 
Each request template is annotated with a regular expression template that it describes.
Since the original dataset is not designed to evaluate generalization to previously unseen request templates, we modified the script provided by the authors to generate a new dataset where the simulation and evaluation requests are generated from disjoint sets of request templates. 
We select 110 regular expressions templates that are each annotated with more than one request template. 
Then, we further remove pairs of regular expression and request templates that are mistakenly paired. 
We end up with 1111 request templates describing these 110 regular expression templates.
We use these templates to generate tuples of requests and regular expressions.
In the end, our dataset consists of 114,503 simulation, 6,429 validation, and 6,429 test data points.
The request templates in the simulation, validation, and test sets are disjoint.  

\paragraph{Simulated Teacher.}
We extend the performance metric $\texttt{perf}$ in \autoref{sec:simulation} to evaluating multiple executions. 
Concretely, given executions $\{ w^{\textrm{inp}}_j, \hat{w}^{\textrm{out}}_j \}_{j = 1}^J$, the metric counts how many pairs where the predicted output word matches the ground-truth: $\sum_{j = 1}^J \mathds{1}\left\{ \hat{w}^{\textrm{out}}_j = w^{\textrm{out}}_j\right\}$. 
We set the threshold $\tau = J$.

\paragraph{Approximate marginal $P_{\piomega}(e \mid s_1)$.} 
The approximate marginal is a uniform distribution over a dataset of (unlabeled) regular expressions. 
These regular expressions are generated using the code provided by \citet{andreas-etal-2018-learning}.\footnote{\url{https://github.com/jacobandreas/l3/blob/master/data/re2/generate.py}}

\begin{algorithm}[t!]
\caption{\mainalg: Learning from Activity Describers via Semi-Supervised Exploration (experimental version).}
\label{alg:practical_adel}
\begin{algorithmic}[1]
\State \textbf{Input}: teacher model $\PP_T(d \mid e)$, approximate marginal $\PP_{\piomega}(e \mid s_1)$, mixing weight $\lambda \in [0, 1]$
\State Initialize policy $\pi_{\theta}: \Scal \times \Dcal \rightarrow \Delta(\Acal)$
\State Initialize policy $\pi_{\beta}: \Scal \times \Dcal \rightarrow \Delta(\Acal)$
\For{$n = 1, 2, \cdots, N$}
\State Word samples $q=(R, s_1, d^{\star}) \sim \PP^{\star}(\cdot)$
\State Agent generates $\hat{e} \sim \PP_{\pibeta}(\cdot \mid s_1, d^{\star})$
\State Teacher generates $\hat{d} \sim \PP_T(\cdot \mid \hat{e})$
\State Agent samples $\tilde{e} \sim \PP_{\piomega}(\cdot \mid s_1) $
\State Compute losses:
\begin{align}
    \mathcal{L}(\theta) &= \sum_{(s, \hat{a}_s) \in \hat{e}} \log \pi_{\theta}(\hat{a}_s \mid s, \hat{d}) \nonumber \\
    \mathcal{L}(\beta) &= \lambda \sum_{(s, \tilde{a}_s) \in \tilde{e}} \log \pi_{\beta}(\tilde a_s \mid s, \hat{d}) + 
    (1 - \lambda) \sum_{(s, \hat a_s) \in \hat{e}} \log \pi_{\beta}(\hat a_s \mid s, \hat{d}) \nonumber 
\end{align}
\State Compute gradients $\nabla\mathcal{L}(\theta)$ and $\nabla\mathcal{L}(\beta)$
\State Use gradient descent to update $\theta$ and $\beta$ with $\nabla\mathcal{L}(\theta)$ and $\nabla\mathcal{L}(\beta)$, respectively
\EndFor
\Return $\pi : s, d \mapsto \argmax_a \pi_{\theta}(a \mid s, d)$
\end{algorithmic}
\end{algorithm}

\section{Practical Implementation of $\mainalg$}
\label{sec:adel_practical}

In our experiments, we employ the following implementation of $\mainalg$ (\autoref{alg:practical_adel}), which learns a policy $\pibeta$ such that $\PP_{\pibeta}(e \mid s_1, d)$ approximates the mixture $\tilde{\PP}(e \mid s_1, d)$ in \autoref{alg:adel}. 
In each episode, we sample an execution $\hat{e}$ using the policy $\pibeta$. 
Then, similar to \autoref{alg:adel}, we ask the teacher $\PP_T$ for a description of $\hat{e}$ and the use the pair $(\hat{e}, \hat{d})$ to update the agent policy $\pitheta$. 
To ensure that $\PP_{\pibeta}$ approximates $\tilde{\PP}$, we draw a sample $
\tilde{e}$ from the approximate marginal $\PP_{\piomega}(e \mid s_1)$ and update $\pibeta$ using a $\lambda$-weighted loss of the log-likelihoods of the two data points $(\tilde{e}, \hat{d})$ and $(\hat{e}, \hat{d})$. 
We only use $(\hat{e}, \hat{d})$ to update the agent policy $\pitheta$.

An alternative (naive) implementation of sampling from the mixture $\tilde{\PP}$ is to first choose a policy between $\piomega$ (with probability $\lambda$) and $\pitheta$ (with probability $1 - \lambda$), and then use this policy to generate an execution. 
Compared to this approach, our implementation has two advantages:
\begin{enumerate}
    \item Sampling from the mixture is simpler: instead of choosing between $\pitheta$ and $\piomega$, we always use $\pibeta$ to generate executions;
    \item More importantly, samples are more diverse: in the naive approach, the samples are either completely request-agnostic (if generated by $\piomega$) or completely request-guided (if generated by $\pitheta$). As a machine learning-based model that learns from a mixture of data generated by $\piomega$ and $\pitheta$, the policy $\pibeta$ can generalize and generate executions that are partially request-agnostic.
\end{enumerate}

\paragraph{Effects of the Annealing Mixing Weight.} We do not anneal the mixing weight $\lambda$ in our experiments. \autoref{tab:anneal} shows the effects of annealing the mixing weight with various settings. 
We find that annealing improves the sample complexity of the agents, i.e. they reach a substantially high success rate in less training episodes. 
But overall, not annealing yields slightly higher final success rates.

\begin{table}[t!]
\centering
\footnotesize
\begin{tabular}{lcccc}
    \toprule
     & \multicolumn{2}{c}{\nav} & \multicolumn{2}{c}{\regex} \\
     \cmidrule(lr){2-3} 
     \cmidrule(lr){4-5} 
     \multicolumn{1}{c}{Anneal $\lambda$ every L steps} & Success rate (\%) $\uparrow$ & Sample complexity $\downarrow$ & Success rate (\%) $\uparrow$ & Sample complexity $\downarrow$  \\  \midrule
     L = 2000 & 31.4 & 304K & 87.7 & 368K \\
     L = 5000 & 32.5 & 384K & 86.4 & 448K \\
     No annealing (final) & 32.0 & 384K & 88.0 & 608K \\
     \bottomrule
\end{tabular}
\caption{Effects of annealing the mixing weight $\lambda$. When annealed, the mixing weight is updated as $\lambda \leftarrow \max(\lambda_\textrm{min}, \lambda \cdot \beta)$, where the annealing rate $\beta = 0.5$ and the minimum mixing rate $\lambda_{\textrm{min}} = 0.1$. Initially, $\lambda$ is set to be 0.5. All results are on validation data. Sample complexity is the number of training episodes required to reach a success rate of at least $c$ ($c = 30\%$ in \nav, and $c = 85\%$ in \regex).}
\label{tab:anneal}
\end{table}

\section{Training details}
\label{app:training}

\paragraph{Reinforcement learning's continuous reward.} In \regex, the continuous reward function is 
\begin{align}
    \frac{|w^{\textrm{out}}| - \texttt{editdistance}\left( \hat{w}^{\textrm{out}}, w^{\textrm{out}} \right)}{|w^{\textrm{out}}|}
\end{align} where $w^{\textrm{out}}$ is the ground-truth output word, $\hat{w}^{\textrm{out}}$ is the predicted output word, \texttt{editdistance}(.,.) is the string edit distance computed by the Python's \href{https://pypi.org/project/editdistance/}{editdistance} module.  

In \nav, the continuous reward function is
\begin{align}
    \frac{\texttt{shortest}\left(s_1, s_g\right) - \texttt{shortest}\left(s_H, s_g\right)}{\texttt{shortest}\left(s_1, s_g\right)}
\end{align} where $s_1$ is the start location, $s_g$ is the goal location, $s_H$ is the agent's final location, and \texttt{shortest}$(.,.)$ is the shortest-path distance between two locations (according to the environment's navigation graph). 

\begin{figure}[t!]
    \centering
    \begin{subfigure}[t]{0.45\linewidth}
        \includegraphics[width=\linewidth]{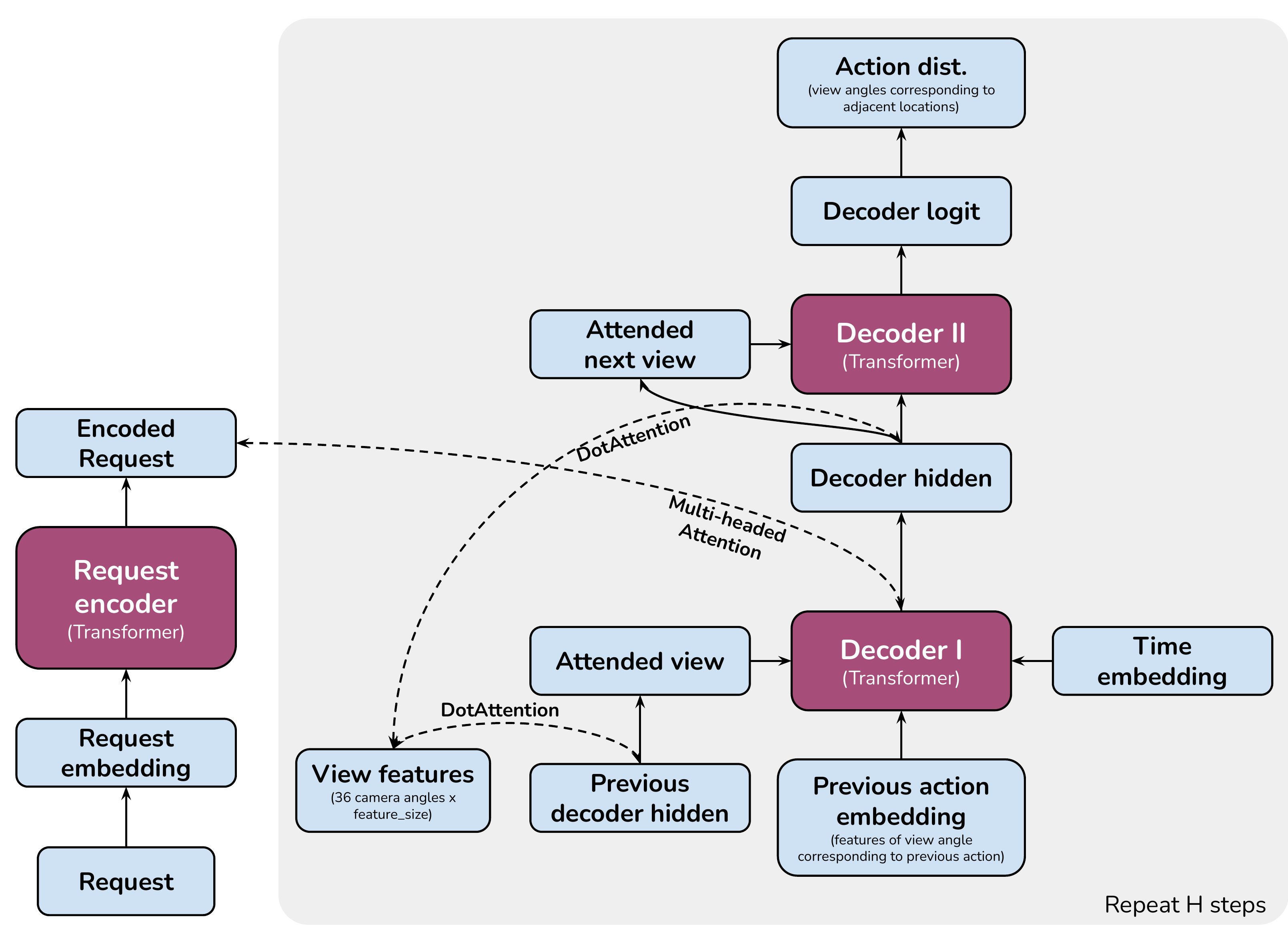}
        \caption{Student model}
    \end{subfigure}
    ~
    \begin{subfigure}[t]{0.45\linewidth}
        \includegraphics[width=\linewidth]{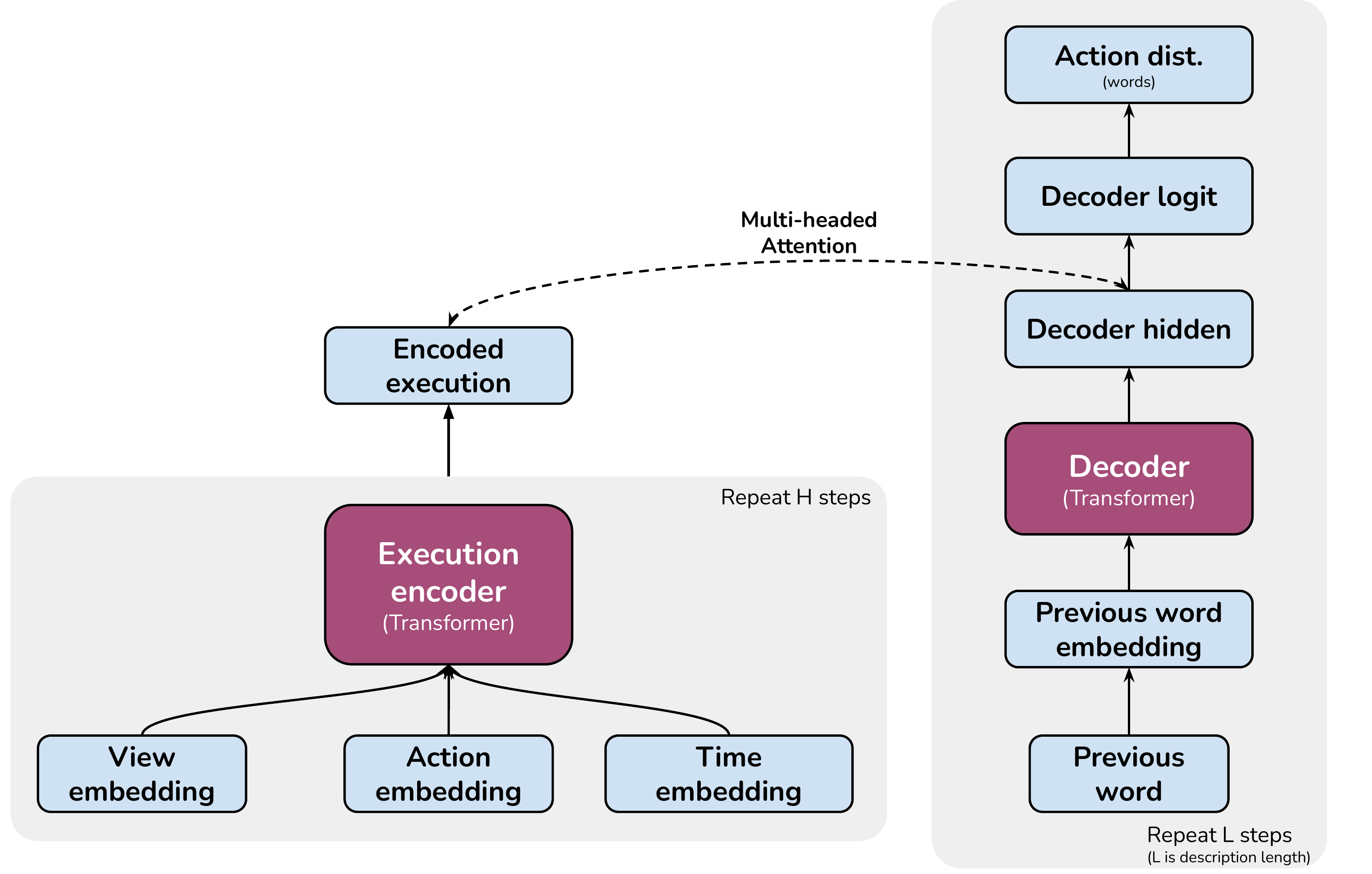}
        \caption{Teacher model}
    \end{subfigure}
    \caption{Student and teacher models in \nav.}
    \label{fig:arch_nav}
\end{figure}

\begin{figure}[t!]
    \centering
    \begin{subfigure}[t]{0.55\linewidth}
        \includegraphics[width=\linewidth]{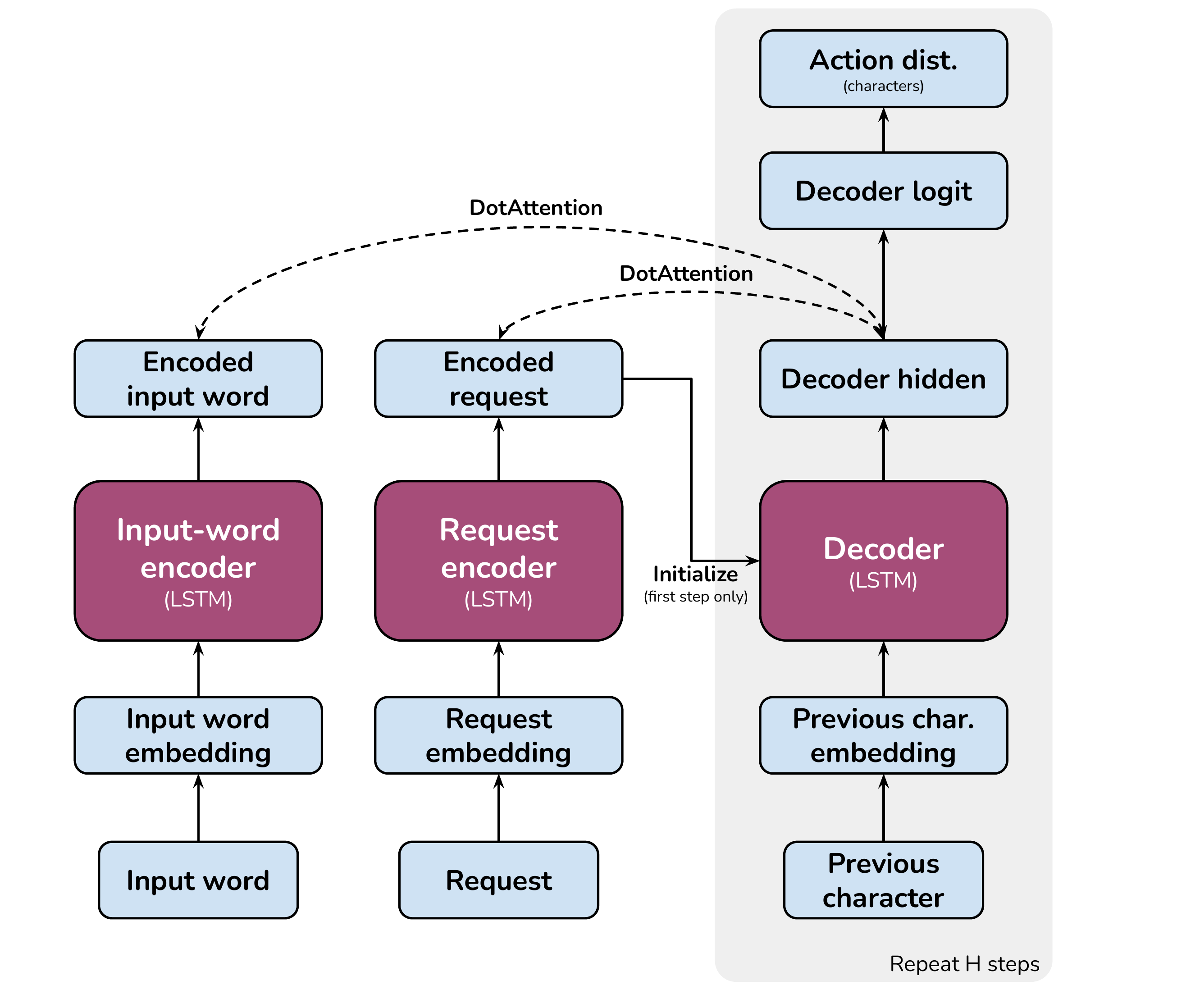}
        \caption{Student model}
    \end{subfigure}
    ~
    \begin{subfigure}[t]{0.35\linewidth}
        \includegraphics[width=\linewidth]{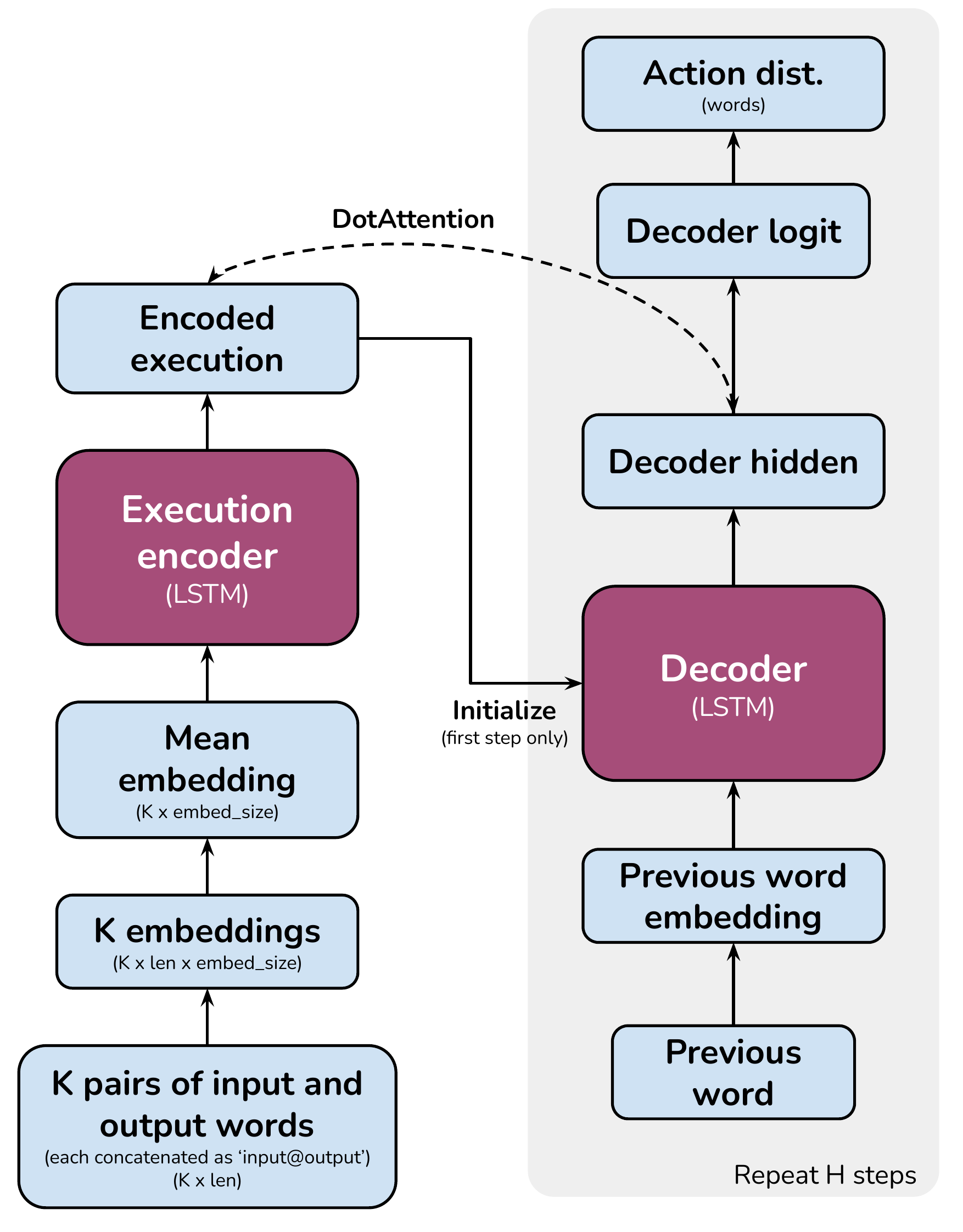}
        \caption{Teacher model}
    \end{subfigure}
    \caption{Student and teacher models in \regex.}
    \label{fig:arch_regex}
\end{figure}

\paragraph{Model architecture.} \autoref{fig:arch_nav} and \autoref{fig:arch_regex} illustrate the architectures of the models that we train in the two problems, respectively.  
For each problem, we describe the architectures of the student policy $\pitheta(a \mid s, d)$ and the teacher's language model $\tilde{\PP}(d \mid e)$. 
All models are encoder-decoder models but the \nav models use Transformer as the recurrent module while \regex models use LSTM. 

\paragraph{Hyperparameters.} Model and training hyperparameters are provided in \autoref{tab:hyper}. Each model is trained on a single NVIDIA V100 GPU, GTX 1080, or Titan X. Training with the \mainalg algorithm takes about 19 hours for \nav and 14 hours for \regex on a machine with an Intel i7-4790K 4.00GHz CPU and a Titan X GPU. 

\begin{table}[t!]
\centering
\scriptsize
\begin{tabular}{lcc}
    \toprule
     \multicolumn{1}{c}{Hyperparameter} & \nav & \regex   \\  \midrule
     \multicolumn{1}{l}{\textbf{Student policy $\pitheta$} and \textbf{Teacher's describer model $\tilde{P}_T$}} \\
     Base architecture & Transformer & LSTM \\ 
     Hidden size & 256 & 512 \\ 
     Number of hidden layers (of each encoder or decoder) & 1 & 1 \\ 
     Request word embedding size & 256 & 128 \\ 
     Character embedding size (for the input and output words) & - & 32 \\
     Time embedding size & 256 & - \\
     Attention heads & 8 & 1 \\
     Observation feature size & 2048 & - \\
     \midrule
     \multicolumn{1}{l}{\textbf{Teacher simulation}} \\ 
     \texttt{perf} metric & STDW \citep{magalhaes2019general} & Number of output words matching ground-truths \\
     Number of samples for approximate pragmatic inference ($|\Dcal_{\textrm{cand}}|$) & 5 & 10 \\
     Threshold ($\tau$)  & 0.5 & $J = 5$ \\
      \midrule
     \multicolumn{1}{l}{\textbf{Training}} \\
     Time horizon ($H$) & 10 & 40 \\
     Batch size & 32 & 32 \\ 
     Learning rate & $10^{-4}$ & $10^{-3}$ \\ 
     Optimizer & Adam & Adam \\ 
     Number of training iterations & 25K & 30K \\
     Mixing weight ($\lambda$, no annealing) & 0.5 & 0.5 \\
     \bottomrule
\end{tabular}
\caption{Hyperparameters for training with the \mainalg algorithm.}
\label{tab:hyper}
\end{table}

\section{Qualitative examples}
\label{app:qual_example}

\begin{figure}[t!]
    \centering
    \begin{subfigure}[t]{\linewidth}
        \centering
        \includegraphics[width=0.7\linewidth]{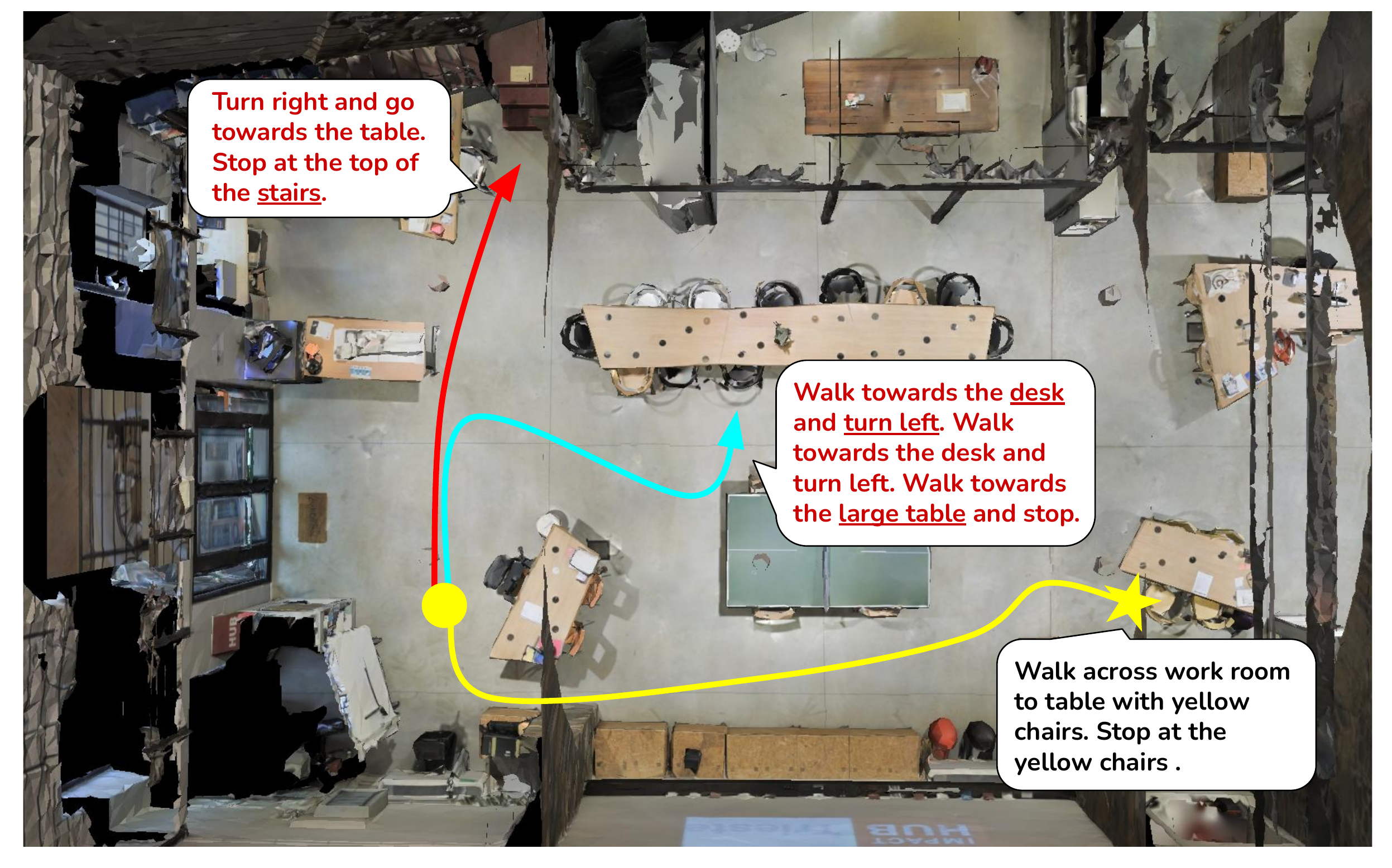}
        \caption{}
    \end{subfigure}
    ~
    \begin{subfigure}[t]{\linewidth}
        \centering
        \includegraphics[width=0.7\linewidth]{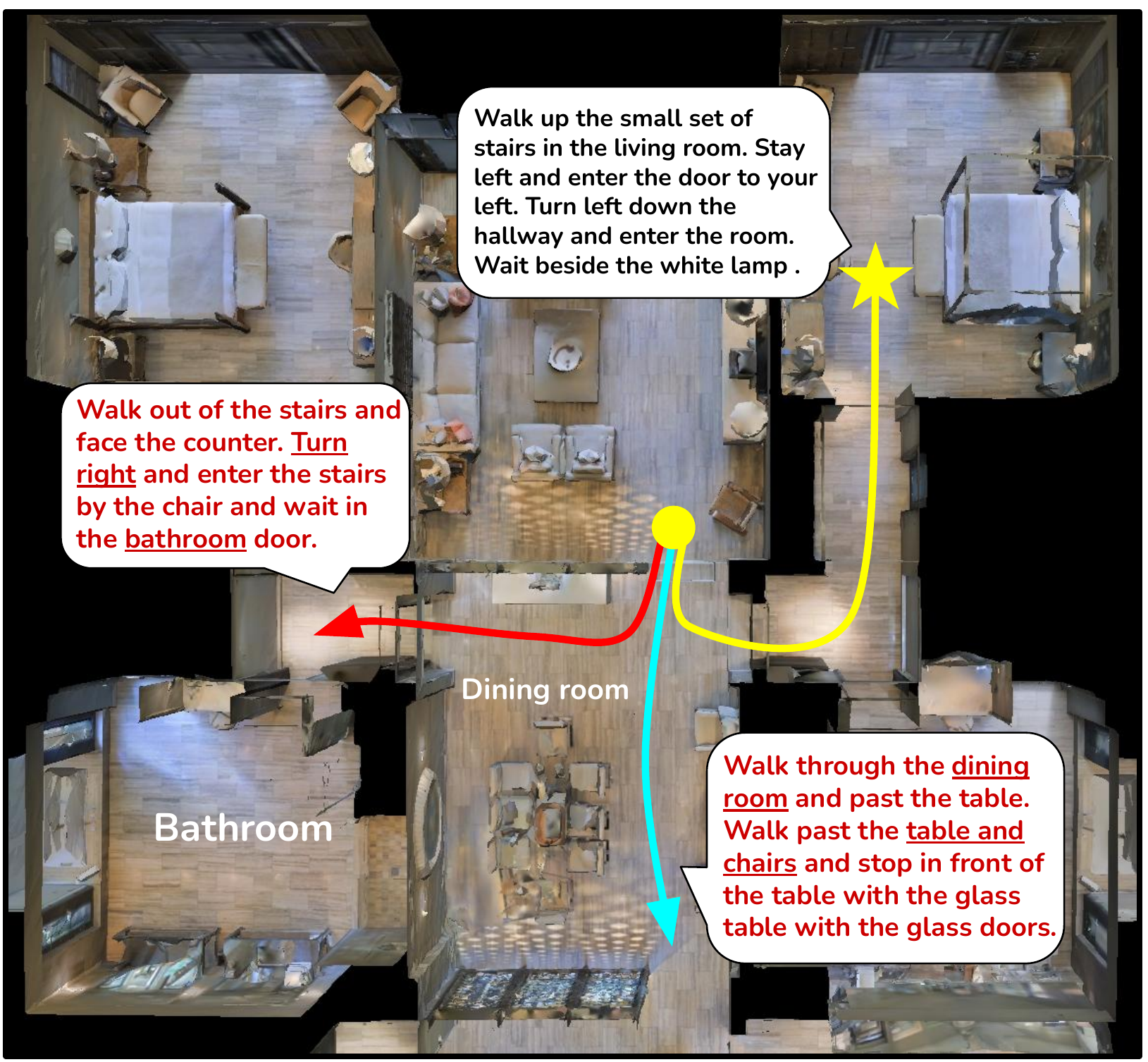}
        \caption{}
    \end{subfigure}
    \caption[hello]{Qualitative examples in the \nav problem. The \textbf{black texts} (no underlines) are the initial requests $d^{\star}$ generated by humans. The \yellowRoute paths are the ground-truth paths implied by the requests. \cyanRoute and \redRoute are some paths are taken by the agent during training. Here, we only show two paths per example. The \textbf{\textcolor{red}{red texts}} are descriptions $\hat{d}$ generated by the teacher's learned (conditional) language model $\tilde{\PP}(d \mid e)$. We show the bird-eye views of the environments for better visualization, but the agent only has access to the first-person panoramic views at its locations.}
    \label{fig:qual_nav}
\end{figure}

\autoref{fig:qual_nav} and \autoref{tab:qual_regex} show the qualitative examples in the \nav and \regex problems, respectively. 

\begin{table}[t!]
\centering
\small
\begin{tabular}{lll}
    \toprule
     Input word & Output word & Description generated by $\tilde{\PP}(d \mid e)$   \\ \midrule
     \textcolor{red}{at}tendant & \textcolor{red}{xj}tendxjt & replace [ a ] and the letter that follows it with an [ x j ] \\
     \textcolor{red}{d}isclaims & \textcolor{red}{e}sclaims & if the word does not begin with a vowel , replace the first two letters with [ e ] \\
     incu\textcolor{red}{l}pating & incu\textcolor{red}{xl}pating & for any instance of [ l ] add a [ x ] before the [ l ] \\ 
     \textcolor{red}{f}lanneling & \textcolor{red}{g}lanneling & change the first letter of the word to [ g ] \\ 
     \textcolor{red}{d}hoti & \textcolor{red}{j}hoti & replaced beginning of word with [ j ] \\
     stuccoing & \textcolor{red}{o}stuccoing & all words get a letter [ o ] put in front \\ 
     reappearance\textcolor{red}{s} & reappearance\textcolor{red}{d} & if the word ends with a consonant , change the consonant to [ d ] \\
     \textcolor{red}{b}i\textcolor{red}{g}o\textcolor{red}{ts} & \textcolor{red}{vy}i\textcolor{red}{vy}o\textcolor{red}{vyvy} & replace each consonant with a [ v y ] \\
     \bottomrule
\end{tabular}
\caption{Qualitative examples in the \regex problem. We show pairs of input and output words and how the teacher's language model $\tilde{\PP}(d \mid e)$ describes the modifications applied to the input words.}
\label{tab:qual_regex}
\end{table}

\end{document}